# Generative Design by Reinforcement Learning: Enhancing the Diversity of Topology Optimization Designs

Seowoo Jang

Department of Electrical and Computer Engineering

Seoul National University

sjang@netlab.snu.ac.kr

Soyoung Yoo

Department of Mechanical Systems Engineering

Sookmyung Women's University

ysy@sm.ac.kr

Namwoo Kang[*]

Department of Mechanical Systems Engineering

Sookmyung Women's University

nwkang@sm.ac.kr

[*]Corresponding author

## Abstract

Generative design refers to computational design methods that can automatically conduct design exploration under constraints defined by designers. Among many approaches, topology optimization-based generative designs aim to explore diverse topology designs, which cannot be represented by conventional parametric design approaches. Recently, data-driven topology optimization research has started to exploit artificial intelligence, such as deep learning or machine learning, to improve the capability of design exploration. This study proposes a reinforcement learning (RL) based generative design process, with reward functions maximizing the diversity of topology designs. We formulate generative design as a sequential problem of finding optimal design parameter combinations in accordance with a given reference design. Proximal Policy Optimization is used as the learning framework, which is demonstrated in the case study of an automotive wheel design problem.  To reduce the heavy computational burden of the wheel topology optimization process required by our RL formulation, we approximate the optimization process with neural networks. With efficient data preprocessing/augmentation and neural architecture, the neural networks achieve a generalized performance and symmetricity-reserving characteristics. We show that RL-based generative design produces a large number of diverse designs within a short inference time by exploiting GPU in a fully automated manner. It is different from the previous approach using CPU which takes much more processing time and involving human intervention.

# 1. Introduction

Generative design refers to computational design methods that can automatically conduct design exploration under constraints defined by designers (Shea et al., 2005; Krish 2011; Singh & Gu, 2012; Oh et al., 2019). It proposes initial designs at the conceptual level and inspires designers to create new designs that have not been previously thought of (Kallioras & Lagaros, 2020).

The parametric representation of design geometry is the most widely used approach to generative design because of parametric computer-aided design (CAD) tools. Recently, generative design has begun to use topology optimization, taking advantage of the ability to represent different topologies in given design spaces (Matejka et al., 2018; Oh et al., 2019; Sun & Ma, 2020). Topology optimization is originally a design optimization methodology to find a single best design, but it can be used as a design exploration methodology through the following approaches: finding local optima, finding a Pareto set by solving a multiobjective (disciplinary) optimization problem, and diversifying the definition of a topology optimization problem. This trend has recently attracted attention as a new design paradigm in the industry (Autodesk, 2021).

In addition, with the advancement of artificial intelligence, various studies have proposed to combine deep learning and topology optimization. Such studies can be categorized as follows: studies that accelerate the iterations of topology optimization (Banga et al., 2018; Lin et al., 2018; Sosnovik & Oseledets, 2019), studies that perform topology optimization without iteration (Cang et al., 2019; Li et al., 2019; Yu et al., 2019; Zhang et al., 2019; Abueidda et al., 2020), studies that enable efficient topology optimization (Guo et al., 2018; Hoyer et al., 2019; Sasaki & Igarashi, 2019), and studies on the application of deep learning to topology optimization-based generative design (Oh et al., 2018; Oh et al., 2019; Yoo et al., 2020; Kallioras & Lagaros, 2020; Sun & Ma, 2020). A detailed introduction to each category is presented in Section 2.

The present study falls into the category of applying deep learning to topology optimization-based generative design. The problem formulation in this study originated from the work by Oh et al. (2019) using a reference design as input to create a topologically optimzed design similar to the reference while improving engineering performance (i.e., low compliance). However, to generate new designs, parameters for design exploration (e.g., weights for multiobjective optimization problem and design conditions) must be determined for each reference design, and the variation of the generated design is greatly influenced by these parameters.

This study proposes a reinforcement learning (RL)-based generative design framework that can enhance the diversity of generated designs through topology optimization. The proposed framework can immediately present a set of optimized design parameters, thereby encouraging the diversity of the designs based on a given reference. This study focuses on developing two neural networks in the proposed RL framework: one that approximates the topology optimization process (*TopOpNet*) and another to find a combination of design parameters leading to maximum diversity (*GDNet)*.

First, we exploit the universal function approximation characteristic of neural networks to perform the highly nonlinear and complex operation of a conventional topology optimization algorithm in more computationally efficient ways. This is carried out by *TopOpNet*, our proposed neural network, that achieves results similar to those of the conventional algorithm with a limited amount of data. It shows even better in the cases when the algorithm fails to maintain symmetric consistency.

Second, on top of the *TopOpNet*, taking advantage of the reduction in computation time, we suggest a novel approach of using a RL framework to define the optimization problem, which we term *GDNet.* There are previous work in the literature that apply RL when designing the engineering structure of an apparatus in a fully automated manner (Li & Malik, 2016; Yonekura & Hattori, 2019; Ha, 2019). In conjunction with virtual simulation based on real-world restrictions, previous work show the feasibility of RL to design structures that meet engineering requirements. In contrast, the present work focuses on applying RL to suggest design candidates as diverse as possible while fulfilling the physical requirements. To achieve this goal, we address the maximum design diversity problem with neural networks as well.

An RL agent finds optimal control/actions through an interaction with an environment, thus, RL typically does not require labelling tasks which are very labor intensive. This work makes use of the previous topology optimization algorithm for the wheel design which requires heavy computations to generate training dataset in a fully automated manner. Then, the neural network is used as a function approximator to construct a well-defined environment the RL-agent can interact with. Likewise, in an automated way, the RL agent is trained to encourage diversity of topologically optimized designs.

Fig. 1 shows the entire process of our work. As shown in the figure, we first run the conventional topology optimization algorithm to generate training data. Then, *TopOpNet* is trained with the pre-generated data to achieve a well-approximated neural model that provides generalized performance to unseen inputs with much less computation time and power. In addition, we propose an RL framework that uses the significantly reduced optimization time of *TopOpNet*. The trained model, *GDNet*, is shown to produce a set of topologically optimized designs with large diversity. We believe that wide diversity in design can help reduce the efforts and ingenious thinking required to obtain generative designs.

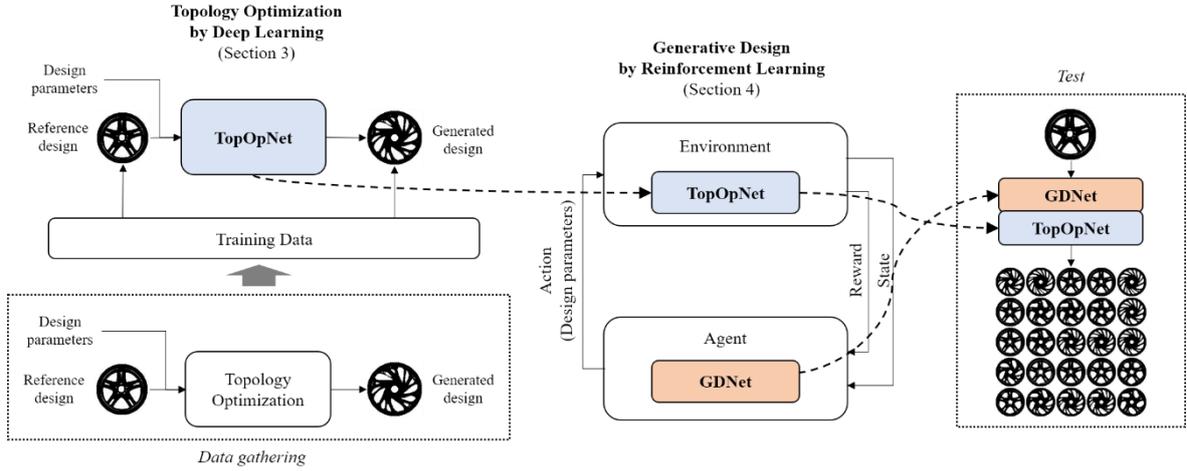

Figure 1. Research framework: reinforcement learning with *TopOpNet* and *GDNet*.

The contributions of this paper are as follows:

- This study proposes a *TopOpNet* architecture with the generalization performance of producing similar-to-algorithm results (and even better in some cases) to unseen inputs. The neural approximation shows improved symmetricity-reserving property due to data augmentation. The results can be achieved in much less time and with much less computational power compared to the iterative topology optimization algorithm. The details on performance evaluations are presented in Section 3.

- This study presents an approach for using RL for the maximum design diversity problem. We find that the problem has a non-trivial solution so that deep learning is a suitable choice for addressing it. The problem is formulated as a sequence of parameter selection stages based on RL rewards. Design choices for the reward function, neural network architecture, and RL framework and dataflow are presented in detail in Section 4.

- This study demonstrates the feasibility of using RL for an automotive wheel design problem as a useful use case, which opens up new applications of RL in the engineering design domain.

The remainder of this paper is organized as follows. In Section 2, we provide related works and preliminaries. In Section 3, we present *TopOpNet* with details on the neural network architecture and evaluation results. Then, we elaborate on the *GDNet*, which generates a set of designs with large diversity while satisfying physical constraints in Section 4. Finally, concluding remarks and future works are discussed in Section 5.

## 2. Related Works

In this section, we provide preliminary background and related work for generative design, topology optimization-based generative design, deep learning-based topology optimization, and RL, which constitute the baselines for our problem.

## 2.1. Generative Design

According to Krish (2011), research on generative design was initiated in the early 1970s by Frazer (Frazer, 2002). Kallioras et al. (2020) stated that the origin of generative design is nature mimicking design algorithms in the 1970s (Lindenmayer, 1975; Meinhardt, 1976). Then, in 1989, with the advent of parametric CAD tools, generative designs were studied in earnest (Shea et al., 2005). Generative design has been applied in various areas of manufacturing, automotive, aerospace and construction, and has been actively studied, especially in the field of construction (Caldas, 2001; Shea et al., 2005; Singh & Gu, 2012).

To define the purpose of generative design, many studies cited Shea et al. (2005): "Generative design systems are aimed at creating new design processes that produce spatially novel yet efficient and buildable designs through exploitation of current computing and manufacturing capabilities". Singh & Gu (2012) explains that the main purpose of generative deign is to explore a larger design space. Generative design can suggest initial designs that designers have not thought of during the concept design stage and provides designers new inspiration (Kallioras & Lagaros, 2020)

Generally, generative design is any digital design model used for design exploration (Singh & Gu, 2012); specifically, Krish (2011) stated that "generative design is a designer driven, parametrically constrained design exploration process, operating on top of history based parametric CAD systems structured to support design as an emergent process". By applying various constraints (geometric viability, manufacturability, cost, and other performance), we can reduce vast design spaces to smaller viable design spaces through parametric representation of design geometry and design result filtering (Krish, 2011). Representative methods used for exploration are cellular automata, genetic algorithms, L-systems, shape grammars, and swarm intelligence (Singh & Gu, 2012).

## 2.2. Topology Optimization-based Generative Design

Topology optimization does not use design geometries (shape or size) as design variables. Instead, it discretizes the entire design space and assigns the material density of each element in the discretized space. The material density in the grid is then used as a design space. This approach can represent various and general topologies, and it is used to find an optimal design for some goals (usually compliance minimization) (Bendsoe & Sigmund, 2013). Topology optimization is a solution to overcome the limitations of parametric generative design, where geometrical parameterizations are insufficient to span a large design space.

However, in terms of design purpose, topology optimization and generative design are conflicting because the former is a design optimization method to find a single best design, whereas the latter aims at design exploration that automatically provides a diverse set of designs that satisfy user-defined design conditions (Kallioras & Lagaros, 2020).

Presented here are basic ideas of how to utilize topology optimization for design exploration, such as generative design instead of finding a single optimal design. The first is to find local optima. One topology optimization problem may have multiple local optima (under the same force and boundary conditions). For example, different optimal topology designs can be obtained in accordance with the initial design, penalization factor, type, and parameters of the filtering method, and termination criteria (number of iterations). By varying these factors, one can find multiple results. The second is to find a Pareto set by solving the multiobjective (disciplinary) optimization problem. For example, by minimizing compliance for two or more load cases, one can find a Pareto set (Kunakote & Bureerat, 2011). Lastly, designers can diversify the definition of the topology optimization problem. Matejka et al. (2018) states that "generative design (topology optimization-based) varies the parameters of the problem definition while parametric design varies parameters of the geometry directly". Designers can define multiple design problems by varying force and boundary conditions, volume fraction, voxel size, materials, and manufacturing constraints. In this manner, it is possible to get as many designs as there are design problems.

In the industry, Autodesk leverages a topology optimization-based generative design approach and offers a practical CAD tool called *Generative Design* (Autodesk, 2021). Autodesk's generative design uses a level set topology optimization and computes combinations of multiple materials and manufacturing methods in parallel via cloud services (Vlah et al., 2020).

In addition to engineering design, user-interface studies that consider interactions with industrial designers are actively conducted for topology optimization-based generative design (Chen et al., 2018; Kazi et al., 2017). Moreover, topological optimization-based generative design combined with deep learning, an area of interest in this study, has evolved into a new type of generative design research (Oh et al., 2019; Sun & Ma, 2020; Kallioras

& Lagaros, 2020), as discussed in the next subsection.

**2.3. Deep Learning-based Topology Optimization**

With machine learning, various data-driven approaches have been studied for topology optimization. For examples, K-means (Liu et al., 2015; Kumar & Suresh, 2019; Qiu et al., 2020), Support Vector Machine (SVM) (Xiao et al., 2019; Lei et al., 2019; Strömberg, 2020), principal component analysis (PCA) (Alaimo et al., 2018; Li et al, 2019; Xiao et al., 2020), Gaussian process (Keshavarzzadeh et al., 2020; Wang et al., 2020), neural networks (Yildiz et al., 2003; Patel & Choi, 2012; Ulu et al., 2016), and random forests (Zhou & Saitou, 2017) are used in the approaches.

In addition, with the recent development of deep learning, research on the application of deep learning to topology optimization has been actively conducted over the past three years. This study focuses on topology optimization studies based on deep learning, and they are classified in accordance with purpose, as follows:

First, the iterative process of topology optimization can be accelerated through deep learning. Topology optimization is performed only on the initial iteration, and the rest of the iterations are replaced by deep learning prediction. Sosnovik & Oseledets (2019) proposed the use of a convolutional encoder–decoder architecture where the inputs are the structure designs and gradient vectors at the intermediate iteration and the outputs are the final optimal structural designs. Banga et al. (2018) extended the same idea to the 3D structural design problem using a 3D encoder–decoder convolutional neural architecture. Lin et al. (2018) modified the U-Net (Ronneberger et al., 2015) architecture, which predicts the final design from an intermediate optimal design without using input gradient vectors to address conductive heat transfer case study.

Second, deep learning can perform topology optimization without iteration. What these studies have in common is that forces and boundary conditions are preprocessed in the form of a matrix that is used as input to a convolutional neural network (CNN). Yu et al. (2019) proposed a two-step process. In the first step, the low-resolution structure is generated using a CNN-based encoder and decoder, and then the high-resolution final structure is generated using a conditionally generative adversarial network (cGAN). Li et al. (2019) used a similar two-step process to the conductive heat transfer structure design problem and used a generative adversarial network (GAN) for the first step and the super-resolution GAN (SRGAN) for the second. Abueidda et al. (2020) predicted the optimal design in just one step, considering material and geometric nonlinearities. Zhang et al. (2019) used tensors as input for U-Net. Rawat & Shen (2019) used Wasserstein GAN (WGAN) to generate 3D structures while using CNN to predict design settings (i.e., volume faction, penalty, and filter radius) to validate the optimal design. Cang et al. (2019) used theory-driven domain knowledge that can help in learning topology optimization data.

Third, a number of research on deep learning has made topology optimization process more efficient. Sasaki & Igarashi (2019) built a deep neural network-based metamodel for the finite element method of topology optimization and applied it to an internal permanent magnet motor design. Guo et al. (2018) reduced the 2D topology design to the latent space using variational autoencoders (VAEs) and style transfer to find the optimal vector in the latent space and reconstruct it to the 2D topology design. This study was applied to a heat conduction design problem. Interestingly, Google tried to solve the topology optimization problem (Hoyer et al., 2019). The CNN generates the structural design, computes compliance through the forward pass, and then calculates the gradient through the backward pass to find the optimal design.

Lastly, some studies combined topology optimization-based generative design and deep learning. This field of study is the focus of our study. Kallioras & Lagaros (2020) created various structural designs using combinations of reduced order models and convolution filters of deep belief networks. Sun & Ma (2020) argued for the limitations of supervised learning's data dependence. They suggested RL-based exploration strategies using ε-greedy policy, upper confidence bound, Thompson sampling, and information-directed sampling. Oh et al. (2018) combined GAN and topology optimization for design exploration. Oh et al. (2019) used a past design (reference) as input data for topology optimization and generated a new design that is similar to the reference while having low compliance. In addition, they proposed iterative design exploration to create a new reference design by learning through boundary equivalent GAN. This study has the advantage of being able to create a real product-like design on the basis of a reference design and has proven its usefulness through the design of car wheels rather than a bracket design used in most studies. Yoo et al. (2020) extended the work of Oh et al. (2019) to provide industrial applications of the 3D wheel CAD/computer-aided engineering process.

**2.4 Reinforcement Learning**

2.4.1. Reinforcement Learning in Engineering Design

RL, which is used for design exploration, seeks learning but has a design optimization component in comparison with supervised learning (Lee et al., 2019). In engineering design research, RL has been used to solve inverse design problems. Cui et al. (2012) applied Q-learning to the ship design optimization problem. Yonekura & Hattori (2019) used double deep Q learning (DQN) to optimize the airfoil angle of attack. Lee et al. (2019) also used double DQN to design microfluidic devices. Sun & Ma (2020) proposed RL-based topology optimization. In addition to structural design, most studies on RL-based inverse design are found in the field of nanophotonics (Dong et al., 2017; Sajedian et al., 2019a; Sajedian et al., 2019b; Badloe et. al. al., 2020; So et al., 2020). RL studies on molecular sequences have also been conducted (Sanchez-Lengeling et al., 2017).

2.4.2. Reinforcement Learning Framework

Unlike supervised learning, an RL agent learns how to act through interactions with an environment. RL is basically an extension of the Markov decision process (MDP), yet does not require explicit models of the interacting environments. Through these interactions, the agent learns optimal actions in terms of long-term expected rewards given the states of the environment. In order to find the long-term relationship between states and actions, the RL agent must fundamentally find a balance between exploration (trying out different actions) and exploitation (making use of already-found optimal actions).

Formally, an RL agent interacts with an environment $\mathcal{E}$, through state $s_t$, reward $r_t$, and action $a_t$ at time $t$. The framework is shown in Fig. 2. The agent then learns to maximize the total accumulated reward and returns $R_t = \sum_k \gamma^k \cdot r_{t+k}$, where $\gamma \in (0, 1]$ is a discount factor reflecting the value of future rewards at present.

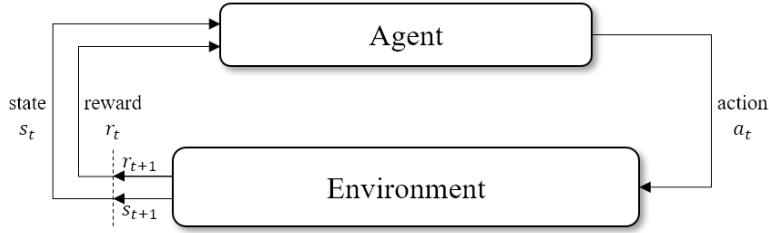

Figure 2. An RL agent interacts with an environment.

There are two categories in RL frameworks depending on how an action is selected given a state. One through value approximation and the other is policy-based.

The action-value function, or Q-function, is the expected reward for selecting an action $a_t$ in state $s_t$, and then follow policy $\pi$. Similarly, the value function is the expected reward from state $s_t$ after following policy $\pi$. The Q-function and value function can be formally expressed as in Eq. (2) and Eq. (3) below, respectively. In the equations, $\theta^Q$ and $\theta^V$ represent parameters for the function approximators for the Q-function and value function, respectively.

$$Q^\pi(s, a; \theta^Q) = \mathbb{E}[R_t | s = s_t, a = a_t] \qquad (2)$$

$$V^\pi(s; \theta^V) = \mathbb{E}[R_t | s = s_t] \qquad (3)$$

Through iterative optimization, the Q or value function is updated and then used to derive the optimal action given the current state, as expressed in the equations below where $P$ is the state transition probability.

$$\pi^*(s_t) = argmax_{a \in A} Q^{\pi^*}(s_t, a; \theta^Q) \qquad (4)$$

$$\pi^*(s_t) = argmax_{a \in A} \sum_{s'} P(s'|s, a) \cdot V^{\pi^*}(s_t; \theta^V) \qquad (5)$$

When using the Q function, a neural network is employed to approximate the value estimation and learn to reduce the value error for the following equations. Examples of the RL framework in this category are DQN and DDQN (Mnih et al., 2013; Mnih et al, 2015; Van et al., 2016).

$$L(\theta^Q) = E\left(r + \gamma \cdot \max_{a_{t+1}} Q(s_{t+1}, a_{t+1}; \theta^Q) - Q(s_t, a_t; \theta^Q)\right)^2 \tag{6}$$

Another category of RL frameworks is the policy gradient methods that work by estimating the policy gradient and updating parameters in the direction of increasing expected rewards $J$. With the policy gradient theorem, the most commonly used estimator has the form of Eq. (7).

$$\theta_{t+1} = \theta_t + \alpha \cdot \nabla \hat{J}(\theta_t) \tag{7}$$

$$\nabla J(\theta) = \mathbb{E}[\nabla log \pi_\theta(a_t|s_t) \cdot \hat{A}_t] \tag{8}$$

where $\pi_\theta$ is a stochastic policy and $\hat{A}_t$ is an estimation of the advantage function at each time step $t$. Here, the advantage is the gap between return and value estimation, which we will cover later in this paper. As in the value function-based RL frameworks, the loss function shown in Eq. (8) plugged into the auto-differentiation software will find optimal parameters $\theta^*$ by minimization. A2C/A3C implements this approach with multiple instances of environments in parallel using the value function approximator as the baseline for the policy update (Mnih et al., 2016).

$$L^{PG}(\theta) = \mathbb{E}[log\, \pi_\theta\,(a_t|s_t) \cdot \hat{A}_t] \tag{9}$$

It is empirically known that using this loss to optimize the parameters in multiple steps leads to a similar trajectory in the parameter space. However, doing so is not well justified, and sometimes fails to learn in the event of large policy updates within a single step (Schulman et al., 2017).

To address this problem, TRPO (Schulman et al., 2015) suggests a surrogate objective for the loss function. The following function is used to impose constraints on the size of the policy update based on the Trust-region theorem (Schulman et al. 2017).

$$\max_\theta \mathbb{E}\left[\frac{\pi_\theta(a_t|s_t)}{\pi_{\theta_{old}}(a_t|s_t)} \cdot \hat{A}_t\right] \tag{10}$$
$$s.t.\; \mathbb{E}[KL\{\pi_{\theta_{old}}(\cdot\,|s_t), \pi_\theta(\cdot\,|s_t)\}] \leq \delta$$

Here, $\theta_{old}$ is the policy parameter before an update and KL is the Kuller-Liebler divergence, which represents the difference between two random distributions. Solving this optimization problem involves multi-step optimization procedures that require extensive computation and time. Therefore, TRPO suggests a relaxed version of the optimization problem as follows.

$$\max_\theta \mathbb{E}\left[\frac{\pi_\theta(a_t|s_t)}{\pi_{\theta_{old}}(a_t|s_t)} \cdot \hat{A}_t - \beta \cdot KL(\pi_{\theta_{old}}(\cdot\,|s_t), \pi_\theta(\cdot\,|s_t))\right] \tag{11}$$

The equation above moves the constraint into the objective function as a penalty with a coefficient $\beta$. The loss function in TRPO is the surrogate function with a penalty. This significantly reduces the complexity of solving the optimization steps of the previous formulation.

However, the relaxed problem above still requires quite a large amount of computation for each step to run. Therefore, PPO (Schulman et al., 2017) suggests a rough approximated version of TRPO, as in the following equation.

$$L^{PPO}(\theta) = \mathbb{E}[\min(r_t(\theta) \cdot \hat{A}_t, clip(r_t(\theta), 1-\epsilon, 1+\epsilon) \cdot \hat{A}_t] \tag{12}$$

$$r_t(\theta) = \frac{\pi_\theta(a_t|s_t)}{\pi_{\theta_{old}}(a_t|s_t)} \tag{13}$$

Here, the *clip* function bounds input within $[1-\epsilon, 1+\epsilon]$. In PPO, instead of the heavy computation for calculating the penalty, the amount of update is bounded to some range, which greatly reduces computation time. The most important effect of the clipping loss is to prevent large destructive policy updates in a single update, which is the empirical problem of A2C.

In conjunction with an A2C style implantation and the generalized advantage estimator (GAE) (Schulman, Morits et al. 2015), PPO has shown stable training performance in a wide range of RL benchmark problems. For this reason, this study uses PPO as our main RL framework in this study.

## 3. Topology Optimization by Deep Learning: *TopOpNet*

In this section, we elaborate on *TopOpNet*. *TopOpNet* approximates the topology optimization algorithm for the wheel designing problem, which requires iterative and heavy matrix manipulations resorting on an extensive amount of computation power and time. To capture the underlying non-linearity of the algorithm, we use a neural network. The proposed neural network architecture shows results similar to the results of the algorithm in much less time. The time required to run the topology optimization algorithm is on the order of 10 min, whereas that of *TopOpNet* is less than 10 ms when inferencing. Owing to the generalization characteristic of deep learning, *TopOpNet* sometimes shows better results in cases where the topology optimization algorithm fails to maintain symmetry.

### 3.1. Wheel Generative Design

Section 2.2 introduces three cases of how topology optimization is used for design exploration. The wheel design problem of Oh et al. (2019) corresponds to the second and third cases. First, it is a multiobjective optimization problem to find a Pareto set by simultaneously solving compliance minimization and similarity maximization. Second, the definition of the problem is diversified by varying load conditions and volume fraction. The combination of these parameters affects design diversity and must be determined for each reference design.

The formulation is

$$
\begin{aligned}
\min \quad & f(\mathbf{x}) = \mathbf{U}^T \mathbf{K}(\mathbf{x}) \mathbf{U} + \lambda \|\mathbf{x}_r - \mathbf{x}\|_1 \\
\text{s.t.} \quad & \frac{V(\mathbf{x})}{V_0} = R \\
& \mathbf{K}\mathbf{U} = \mathbf{F} \\
& 0 \leq x_e \leq 1, \; e = 1, \ldots, N_e.
\end{aligned}
\tag{1}
$$

The objective is to minimize compliance, $\mathbf{U}^T\mathbf{K}(\mathbf{x})\mathbf{U}$, and the distance between the target reference design and the generated design, $\|\mathbf{x}_r - \mathbf{x}\|_1$. The design variables used are: $\mathbf{U}$ as the displacement vector, $\mathbf{K}$ as the global stiffness matrix, $\mathbf{F}$ as the force vector, and $\mathbf{x}$ as the density vector of the element. If $x$ is 1, the element is completely full, and if $x$ is 0, the element becomes empty. The L1 norm is used for the distance between designs, and similarity weight $\lambda$ controls the tradeoff between compliance and similarity to the reference design. As an equality constraint, $R$ is the volume fraction, where $V(x)$ is the volume of the generated design, and $V_0$ is the volume of the reference design.

To create various designs, designers can change the design parameters used in Eq. (1). For example, after discretizing the similarity weight $\lambda$ and changing from small to large, designers can create designs that are similar to the reference design, as well as designs that are very different from the reference design.

This study uses the wheel design problem of Oh et al. (2019) to demonstrate the proposed RL framework. We chose two design parameters: force ratio (i.e., the ratio between normal force and shear force) and similarity weight. The reason for choosing these two is that they have the greatest influence on design diversity which has been demonstrated through various experiments in the literature. In previous studies, the design parameter values were chosen at equal intervals between the minimum and maximum values, as shown in Fig. 3. The purpose of our study is to select optimal design parameter combinations according to reference designs that lead to a set of generated designs with as large a diversity score as possible. We assume that the optimal design parameter combinations depend on the reference design.

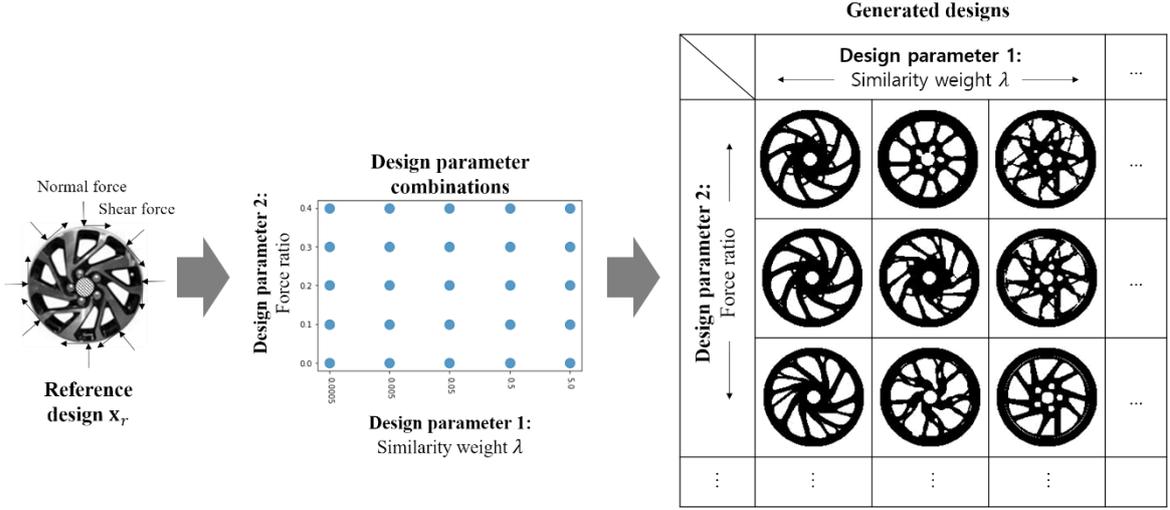

Figure 3. Generative wheel design problem

### 3.2 Training and validating data

A total of 639 reference images are used. These images are obtained using web crawling on a wheel selling website, such as Tire Rack (Tire Rack, 2018). We divide the reference images into two sets: 500 for training and the rest for validation. We randomly split the dataset when evaluating the performance. The training set is used for training the network and the validating set for validating the training process and for testing the generalization performance of the algorithm after the training process.

As discussed in Section 2.1, two design parameters are chosen: similarity weight and force ratio. From now on, we will name similarity weight and force ratio $c_1$ and $c_2$, respectively. The range of the parameters $c_1$ and $c_2$ are then fitted to the problem that we are addressing, the topology optimization of the wheel design. As a result, $c_1 \in [0.0005, 5.0]$ and $c_2 \in [0.0, 0.4]$. These ranges are chosen experimentally from (Oh et al., 2019). In order to generate the training and validating datasets, we sample 11 points in the parameter space at equal distances. We sample $c_1$ in log-scale, whereas linear scale is used for $c_2$. Experiments show that when $c_1$ is controlled on a log-scale, the similarity to the reference design changes remarkably (Oh et al., 2019). Therefore, in total, for one reference image, we get 121 combinations of parameters.

We then derive the 121 topology-optimized images from one reference image with the topology optimization algorithm. Hence, the inputs are a reference image $I_{ref}$, $c_1$, and $c_2$, and the output is a topology-optimized image $I_{opt}$. Thereby, $I_{opt} = f(I_{ref}, c_1, c_2)$. With the pre-calculated datasets $(I_{ref}, c_1, c_2, I_{opt})$, we approximate the function $f$ with a deep neural network.

### 3.3 Network Architectures

We propose a neural network architecture for the *TopOpNet* problem. As discussed, the inputs of the network are a reference image and the two optimization parameters $(I_{ref}, c_1, c_2)$, and the output is the optimized image $I_{opt}$. The size of the image is regularized to 128 × 128 with a single channel, and the parameters are simply two floating numbers. The following methodology is devised to combine an image with two parameters.

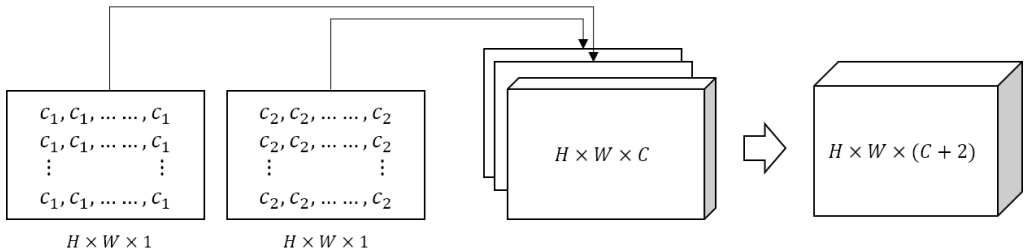

Figure 4. Parameter concatenation layer.

Fig. 4 illustrates the parameter concatenating operation that we propose in this study. Given a tensor (matrix) with shape $H \times W \times C$ where H, W, and C are height, width, and channel respectively, we repeat $c_1$ and $c_2$ to generate the same shape in height and width, only different in channel dimension. Then, we concatenate the two parameter layers on the tensor resulting in an increment on the channel dimension. We perform the parameter concatenation operations at the beginning of each layer to pass the information of the parameters.

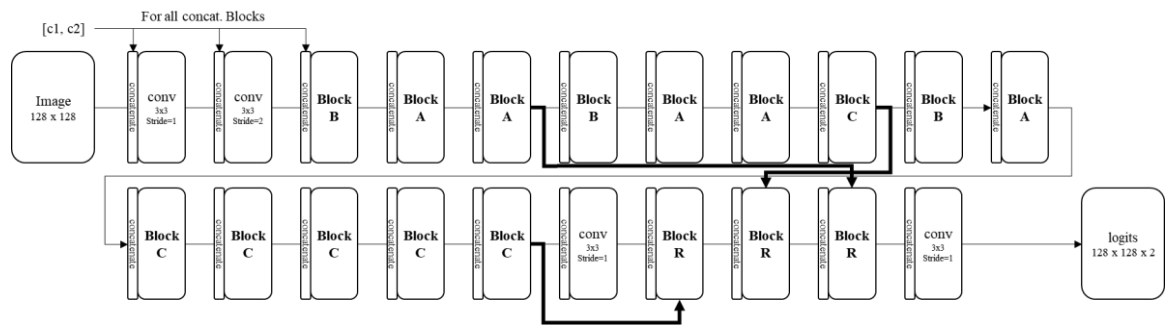

Figure 5. Overall architecture of *TopOpNet*. The model is with ASPP.

The network architecture that we propose is based on the Inception (Szegedy et al., 2017) and Fully convolutional networks (FCN) (Long et al., 2015) architectures with Atrous spatial pyramid pooling (ASPP). Fig. 5 shows the entire architecture of our first proposed model. The network architecture consists of mainly two sets of blocks: encoding blocks, and decoding blocks. The encoding part is a sequence of the blocks A, B, and C, each of which has its own inner network architecture, as shown in Fig. 6. Given an input tensor, the encoding part decreases the height and width of the input tensor by 8. The decoding part consists of three R blocks, each of which doubles the height and width of the input tensor. Thus, our architecture extracts feature from the input tensor by 1/8 and then reconstructs a topologically optimized design with the decoding part.

In order to help preserve the low-level features extracted within the encoding part, we also add three skip layers to pass the features to the decoding part. The skip layers are depicted with bold arrows in Fig. 5. Note that at the beginning of each block, we adopt a concatenated layer to incorporate the parameters $c_1$ and $c_2$.

Fig. 6 shows the architecture of blocks A, B, C, and R. Note that only block B receives the stride as input, which is used to decrease the size of the feature map. By sequentially applying block C, we can construct an ASPP by increasing the dilation size. In block C, we split the feature map into two sets and apply different dilation sizes to capture features with different receptive field sizes. The total number of trainable parameters in *TopOpNet* are approximately 14 million, which is less than the network models used for semantic segmentation in ResNet (He et al., 2016), FCN, and Inception.

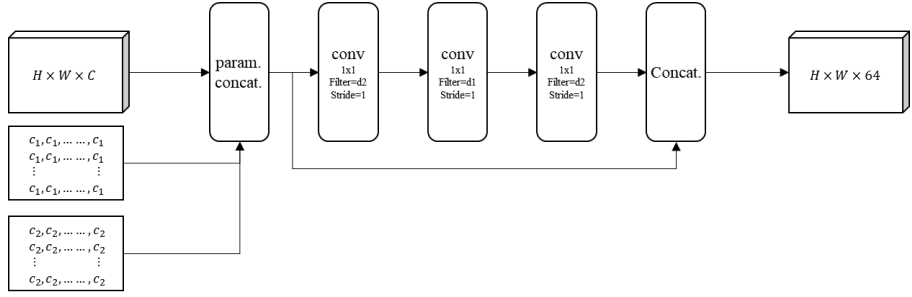

(a) Block A

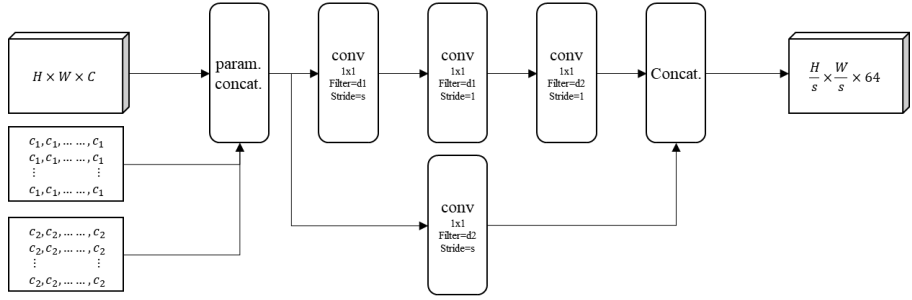

(b) Block B

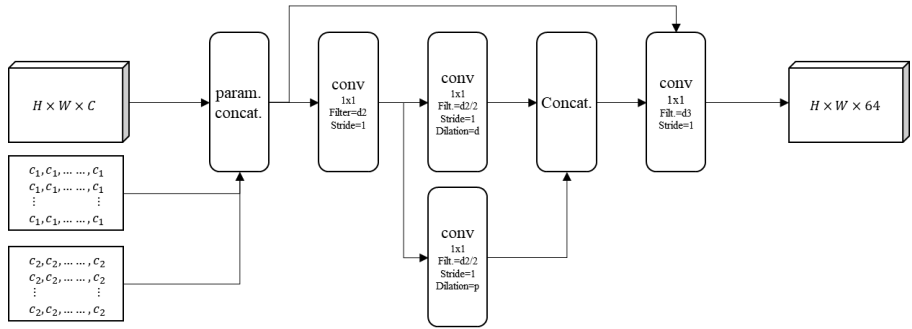

(c) Block C

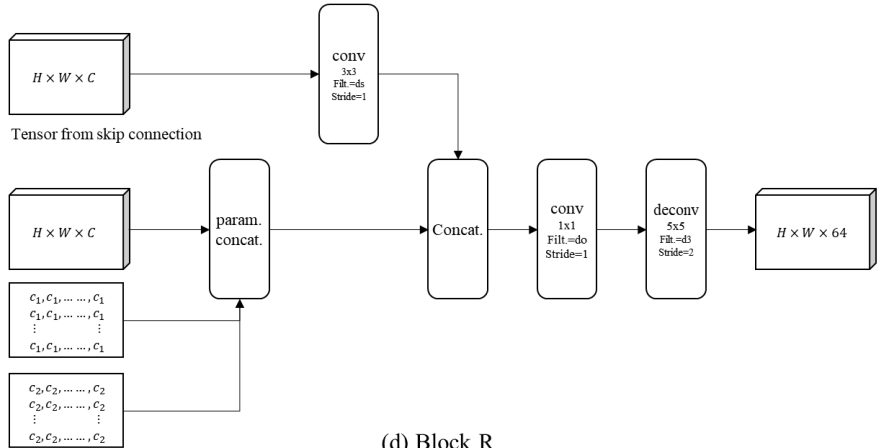

(d) Block R

Figure 6. Blocks in *TopOpNet*.

### 3.4 Training Details and Data Augmentation

For training, we use the Adam optimizer (Kingma & Ba, 2017) with a learning rate of 0.0001. The size of the batch for training is 16; however, a larger batch is preferable if more memory is available on the GPU. As mentioned previously, we randomly split 639 reference images into 500 training and 139 validating/testing images. We also apply weight regularization on the trainable variables with a coefficient of 0.0001. The categorical cross-entropy loss is used since the output image have binary value indicating whether the pixel is filled or not, thus, the problem is binary classification.

In addition, for the generalization performance of the network, we augment the input reference image in two ways. First, Gaussian noise sampled from $\mathcal{N}(0, 0.002)$ is added to each pixel to model pixel-level noise. This helps the neural network learn to recover from pixel-level noise in the input images. And then, the reference and label images are rotated with the same angle. The rotation angle is uniformly sampled from a set [0, 359]. This augmentation makes neural network to learn to preserve symmetricity of the images. Both the augmentations are applied on run-time, i.e., every time a new batch of training data is retrieved, the augmentations are applied.

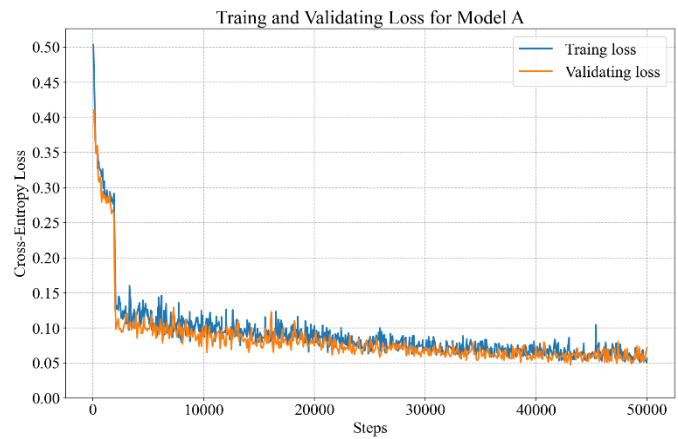

Figure 7. Training and validating loss during training.

As can be seen in Fig. 7, the training and validating losses show a similar trend, which indicates that the training is performed well without overfitting. Here, a training step consists of one forward and back propagation with one batch of training data.

### 3.5 Experiment Results: Visualization, Evaluation, and Comparison

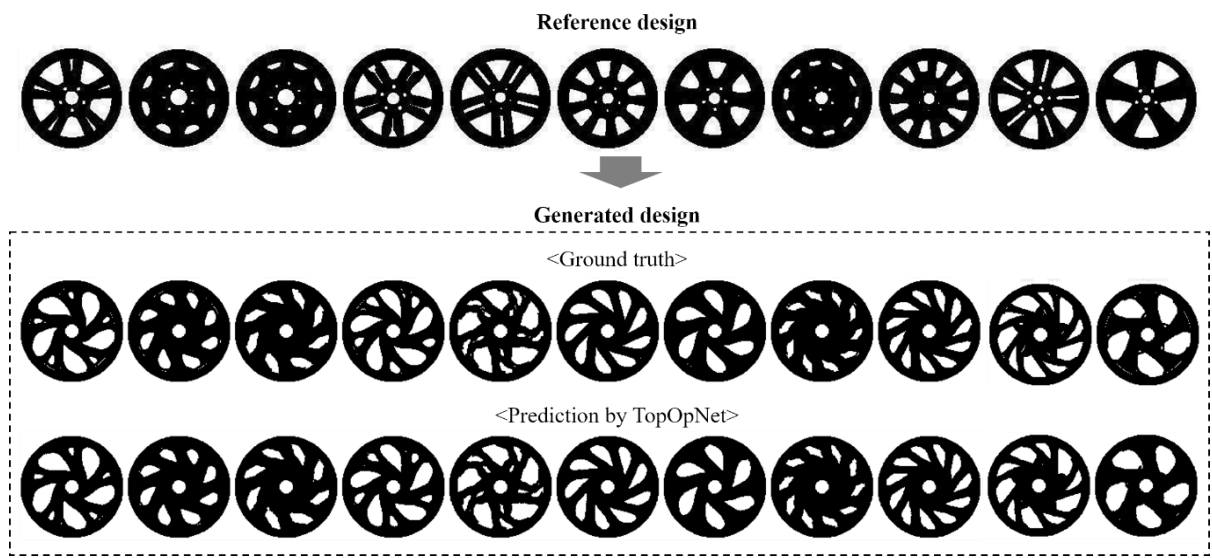

Figure 8. *TopOpNet* results, (top) reference design, (middle) ground truth, (bottom) prediction.

Fig. 8 shows the prediction results of *TopOpNet*. As can be seen in the figure, the neural network approximates the topology optimization algorithm well in most cases. In some cases, the approximation is better, as shown in

Fig. 9. The conventional topology optimization algorithm involves many matrix manipulations such as calculating inverse matrix and determinants which sometimes are not feasible to get, thus, the algorithm has exception handling logics within it. These sometimes leads to the break in the symmetricity in our wheel design. The wheel designs usually have rotational symmetry in nature. For this reason, the rotational augmentations on the training data result in better symmetry preservation. Gao, L., et al (2019) and Li, J., et. al. in their work, SDM-net and GRASS respectively, explicitly include symmetric reserving mechanisms when generating 3D shapes of objects. Since wheel design in our problem is 2D and the designs are all aligned at the center and with the same radius, rotational augmentation effectively reduces asymmetry.

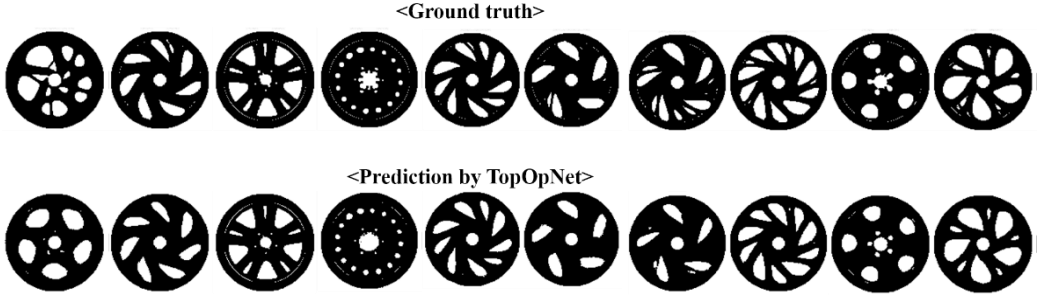

Figure 9. *TopOpNet* better results, (top) ground truth, (bottom) prediction.

Table 1 shows the comparison of the performance of *TopOpNet* with two baseline architectures selected in previous research. Cross validation is performed with 10 random samplings, and the mean and standard deviation are calculated over these 10 tests. For the baseline, we use the architectures of Abueidda et al. (2020) and Lin et al. (2018); such architectures have a single step. We modify the input layer of the two baseline models to fit for our design problem. We also try applying the two-step model of Li et al. (2019), but it does not converge in our problem.

Here, cross entropy, pixel difference, intersection over union (IoU), and compliance metrics are compared. IoU is a metric mainly used in the field of object detection. It can measure how similar the generated designs are. In structural design, even small design differences can remarkably change compliance; thus, comparing the compliance of the generated design with ground truth is crucial. For the compliance metric, we calculate the mean absolute percentage error (MAPE) and use the median of test data, not the mean, because outliers occasionally cause huge errors, which are reflected in the mean.

Table 1. Performance comparison of models.

| Models | Size (#params) | Cross Entropy | | Pixel Difference | | IoU | | Compliance (MAPE) | |
|---|---|---|---|---|---|---|---|---|---|
| | | Mean | Std | Mean | Std | Mean | Std | Mean | Std |
| TopOpNet | 14,630,412 | 0.0341 | 0.0101 | 0.0144 | 0.0050 | 0.9718 | 0.0090 | 0.8851 | 0.2341 |
| Abueidda et al. (2020) | 8,300,248 | 0.0333 | 0.0041 | 0.0118 | 0.0009 | 0.9766 | 0.0016 | 2.7634 | 0.8503 |
| Lin et al. (2018) | 197,906 | 0.0411 | 0.0031 | 0.0167 | 0.0012 | 0.9671 | 0.0024 | 3.5796 | 1.4145 |

For cross entropy, pixel difference, and IoU, Abueidda et al. (2020) shows the best performance among the models, but the results are not significantly different from *TopOpNet*. However, for compliance, *TopOpNet* is significantly better than the rest of the models (i.e., $p<0.001$).

In this section, we propose the *TopOpNet* architecture to approximate the topology optimization algorithm. In general, the topology optimization algorithm is a CPU-intensive job that requires extensive matrix manipulation. The algorithm approximately takes the order of seconds for simple inputs. With our proposed approach and network architectures, the job can be run in less than 1 ms on our NVidia GeForce 1080 Ti GPU. The huge reduction in time is exploited in the next task we address, that is, finding the maximum diversity in designs with RL.

## 4. Generative Design by Reinforcement Learning: *GDNet*

In this section, we address the problem of generating a set of diverse wheel designs using an RL approach on top of *TopOpNet* as an approximated function for topology optimization. In other words, the goal is to propose a set of parameters $\{c_1^i\}, \{c_2^i\}, i \in \{1, 2, ..., n\}$ producing a set of new designs with maximum diversity for an input design (a reference design to generate the variations from).

We first describe the detailed formulation of the problem. Then, the detailed neural architectures and training methods are presented, followed by evaluation results. In the formulation, PPO (Schulman et al., 2017) is used for the RL framework due to its robustness in hyperparameter settings, thus showing good performance in many benchmark RL tasks. Then, the performance evaluation results of our RL-based approach, *GDNet* with equdistance, greedy, and exhaustive search methods, are presented. Typically, RL approaches require various trial-and-error experiences to find optimal actions or controls, and we address this issue by reducing the processing time for applying topology optimization through *TopOpNet*. In contrast to other methods, our *GDNet* can generalize previously unseen input and thus requires single inferencing time to produce optimal actions without searching the entire search space.

### 4.1 Problem Formulation

Fig. 10 depicts the RL workflow of *GDNet*, showing the interaction between an RL agent and an environment. The environment includes the topology-optimization module, *TopOpNet*. Training an RL agent typically requires multiple instances of the environment to gather additional generalized and time-uncorrelated experiences; thus, we use the environment with multiple *TopOpNet* instances within it. The *TopOpNet* module takes a $128 \times 128$ image and generates a topologically optimized image with the same size. Meanwhile, it uses parameters $c_1$ and $c_2$. Thus, the *TopOpNet* module generates a single image, given an input image and a parameter set. As for the parameter set, two sequences of parameters, $\{c_1^i\}$, $\{c_2^i\}, i \in \{1, 2, ..., n\}$, consist of total actions in a single episode. In our problem formulation, $n$ is set to 5, and all combinations of the two sequences are used for topology optimization parameters. Thus, a total of 25 variations are generated from the single reference design at the end of an episode.

To simplify the problem without loss of generality and to compare the results of our *GDNet* with other approaches, we split each parameter space, $c_1$ and $c_2$ spaces, into 11 discrete values. This split leads to discrete action space in our RL formulation. Addressing the problem with continuous action space is possible because *TopOpNet* can infer optimized results with $c_1$ and $c_2$ values in continuous domain. However, an RL agent takes much more computation resources and time to learn the *GDNet* problem. Thus, we stick to discrete action space in this work and leave expanding to continuous action space and exploring continuous domains in an efficient way to future work.

The generated variations are used to calculate a reward value, representing diversity, and they are passed to the RL agent as a state. The RL agent uses the reward and the state as inputs and infers an action that has two values, $c_1$ and $c_2$. The actions are then passed into the environment for the next step. As previously stated, an episode consists of five steps in our formulation. Therefore, at the end of an episode, the diversity of the 25 variants is evaluated as the objective to maximize.

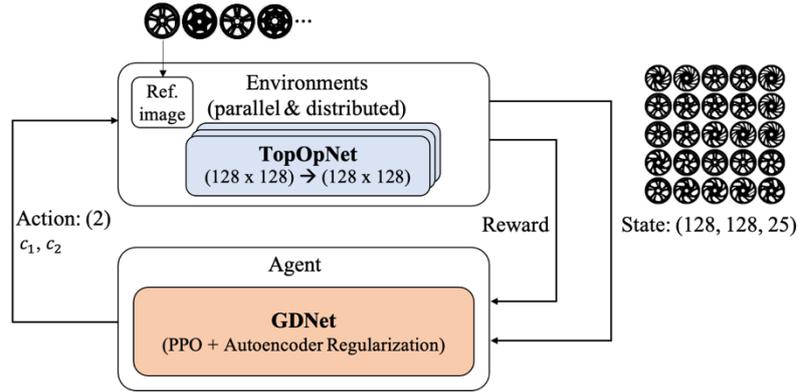

Figure 10. RL pipeline of *GDNet* problem.

Fig. 11 shows the interaction in detail. In the figure, the upper left squares in 5 × 5 grids with numbers show the positions of the resulting images for each time step in an episode. Each axis corresponds to parameters $c_1$ and $c_2$ sequences. Thus, the grid has 25 generated images. The initial state of the environment is 25 duplicated input images (reference design) in the grids. In the figure, the positions in the grids with input image are shown in blank. Then, at each time step, an action of $(c_1, c_2)$ is generated from the agent and passed on to the environment. Then, the environment produces new designs on top of the previously generated ones. The newly stacked images in the grids are the next state, and a diversity value on the state is then calculated as a reward value. The new state and reward are passed to the RL agent. This process continues until an episode ends. This iteration is depicted in the figure as time step evolves.

Our objective is to maximize the diversity of the variants at the end of an episode when all 25 designs are generated. In the middle of an episode, the state transits to the next on the basis of a previous state and action. To achieve maximum diversity at the end, actions leading to local suboptimal results should be selected in the middle. In other words, at each time step, actions must be chosen in consideration of the accumulated reward at the end of an episode, that is, not just considering short-term rewards. To verify the effect of considering long-term expected rewards, we compare the results of our *GDNet* with a greedy algorithm.

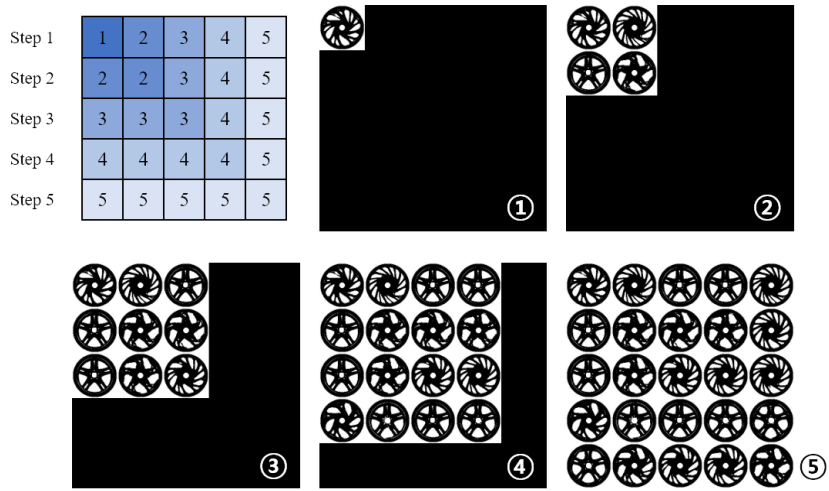

Figure 11. Sequential actions and states

To verify whether our maximum diversity problem has a nonlinear property and does not have a trivial solution, we search the optimal parameter sets leading to the maximum diversity with exhaustive search for numerous reference designs. Here, a trivial solution would be one that can be applied to all reference. Fig. 12 shows the optimal actions for two reference images, and the overlap of all optimal actions for several randomly selected references. As shown in the figure, in some cases, parameters other than the ones at the edges are optimal (middle 0.12 for $c_2$), which means the solutions are not trivial. The overlapping graph (right) shows that the answers have outstanding tendencies but simultaneously have different configurations for each reference input image.

Moreover, we must extract features from an image to derive maximum diversity parameters. The relations among them are nonlinear and not trivial to find with a hand-crafted algorithm.

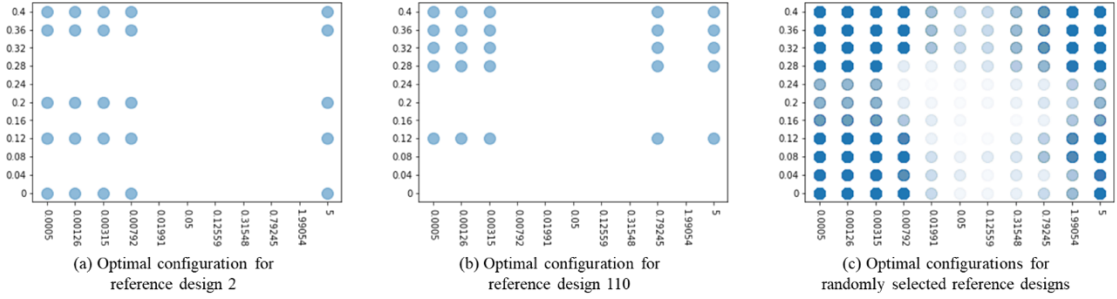

Figure 12. The optimal actions with exhaustive search.

The reward function must capture the diversity of the generated outputs. In this study, we explore two metrics for the diversity: pixel difference and structural dissimilarity (Wang et al., 2004). The pixel difference (pix-diff) is basically the Euclidean distance between two images. Let $x_i$ represent the pixel value of an image $i$. Then, the sum of the L2 distance of $x_i$ from all other images $j$ (pixel-diff value) can be calculated, as in Eq. (14). Then, the average of the L2 norm of the pixel-diff value for all images is as shown in Eq. (15). As it can be derived in the equation, the final value is the difference between the square mean of all values on that pixel subtracted by the square of all means. According to the Cauchy–Schwarz inequality, the pixel difference on the pixel for all images is larger than 0. In Eq. (15), ½ accounts for the duplicated summation for the $(i,j)$ and $(j,i)$ pair. For simplicity of calculation, the pixel difference metric requires less time for its calculation, which is a suitable characteristic for RL. The pixel difference metric is calculated for all pixels in the images and then added.

As discussed earlier, our wheel designs are aligned at the center of images with the same size of $128 \times 128$, and all have the same radius. And each pixel in image represents the existence of material at the position and the existence influence the distribution of forces applied to wheel directly. Therefore, comparing the existence pixel by pixel reflects the performance of designs.

$$\begin{aligned} L_i &= \frac{1}{N}\sum_j (x_j - x_i)^2 \\ &= \frac{1}{N}\sum_j (x_j^2 - 2x_j x_i + x_i^2) \\ &= \frac{1}{N}\sum_j x_j^2 - \frac{2}{N} x_i \cdot \sum_j x_j + x_i^2 \end{aligned} \quad (14)$$

$$\begin{aligned} L &= \frac{1}{2N}\sum_i L_i \\ &= \frac{1}{2N}\sum_i \left(\frac{1}{N}\sum_j x_j^2 - \frac{2}{N}\cdot \sum_j x_j \cdot x_i + x_i^2\right) \\ &= \frac{1}{2N}\left(N \cdot \frac{1}{N}\sum_j x_j^2 - 2 \cdot \frac{1}{N}\sum_j x_j \cdot \sum_j x_j + \sum_j x_j^2\right) \\ &= \frac{1}{2}\left(\frac{1}{N}\sum_j x_j^2 - 2 \cdot \left(\sum_j x_j\right)^2 + \frac{1}{N}\sum_j x_j^2\right) \\ &= \frac{1}{N}\sum_j x_j^2 - \left(\sum_j x_j\right)^2 \end{aligned} \quad (15)$$

By contrast, structure dissimilarity (DSIM) takes the windows of pixels $N_w \times N_w$ and calculates dissimilarity on the windows. Eq. (17) shows the definition of the DSIM. In the equation, SSIM is the structural similarity, which represents the similarity of the two grids $x$ and $y$. Here, $\mu_x, \mu_y$ are the averages, $\sigma_x, \sigma_y$ are the variances, and $\sigma_{xy}$ is the covariance of $x$ and $y$. $k_1$ and $k_2$ are constant values for stabilizing the division with a weak denominator. For our calculations, we use $N_w = 11$. In this case, the total reward is the average of the DSIM value for all pairs ($N$ pairs) of the generated image, as shown in Eq. (18). Qualitatively speaking, DSIM divides images into sliding windows and calculates dissimilarity with statistically representative values, mean, and standard deviation of values in the windows.

$$SSIM(x,y) = \frac{(2\mu_x \mu_y + k_1) \times (2\sigma_{xy} + k_2)}{(\mu_x^2 + \mu_y^2 + k1) \times (\sigma_x^2 + \sigma_y^2 + k_2)} \quad (16)$$

$$DSIM(x,y) = \frac{1 - SSIM(x,y)}{2} \quad (17)$$

$$L = \frac{1}{N}\sum_{i,j} DSIM(i,j) \qquad (18)$$

These two rewarding metrics intuitively make sense but whether which one is better for our problem remains unclear. Thus, two RL agents are trained with each rewarding metric, and the final diversity results are evaluated with both to identify which rewarding metric is more efficient. We show the results later in this paper.

Pixel difference and DSIM may vary considerably if each design's position (offset) slightly changes for the same set of designs. Designs in our dataset are perfectly aligned so that the center and the radius of the designs are always identical. Thus, using these two reward functions has no offset error issue.

We elaborate the detailed state, action, reward, state transitions, and the interaction between the RL agent and the environment. Next, we focus the architecture of neural networks and the framework of PPO and provide training details.

### 4.2 Structure of GDNet and Training

As discussed earlier, we use PPO as our RL framework for its robustness in training stability and performance. However, to boost the feature extracting capability of PPO, we add an additional branch, i.e., VAE regularizer. Fig. 13 depicts the *GDNet* architecture, which is composed of an encoder, a policy, a value, and a decoder network. As stated, inputs to our *GDNet* are stacked images or a tensor of shape $(128 \times 128 \times 25)$. Then, the agent generates a probability distribution on $c_1$ and $c_2$. Given that we stick to discrete action space, the probability distributions on 11 discrete values for each parameter are the outputs of the policy network; therefore, the shape of the probabilistic policy is $(2 \times 11)$. Moreover, the value network keeps estimating the value of the given states by generating a scalar value for a state. As stated in the PPO paper (Schulman et al., 2017), we use a shared embedding architecture in which an encoder extracts embedding features (embedding) from the input tensor; then, the embedding vector, which has a shape of 2048, is fed into the policy and the value networks.

In addition to the basic network blocks, we adopt a decoding network to reconstruct the original inputs from the embedding vector. The decoding networks are inspired by the VAE architecture (Kingma & Welling, 2013), which extracts features from the input and reconstructs the original inputs from the embedding vector. In this manner, VAE compresses the input data within an embedding vector. In our architecture, the encoding and decoding blocks correspond to the VAE, which enhances the feature extraction of the entire architecture. In other words, we use VAE as a regularizer for the network to avoid overfitting. The entire architecture of our PPO agent is shown in Fig. 13.

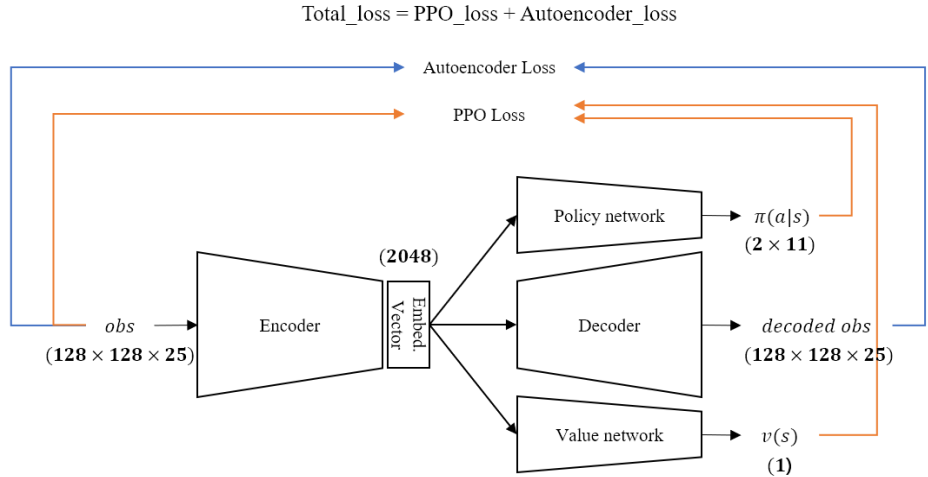

Figure 13. The PPO agent with the VAE regularizer.

An encoder, policy, value, and decoder network for our agent can be constructed with various neural architectures. We use a simple CNN with convolutional layers for the encoder and deconvolution layers for the decoder. For the policy and the value networks, we use simple multilayer perceptron layers.

The *GDNet*, our customized PPO framework, is then trained with 500 input reference images (original designs), and the performance of it is evaluated with the remaining 139 reference images as in the case of *TopOpNet*. For fair comparison between rewarding metrics, we use a fixed number of training steps of 5,000, which corresponds to 1,000 episodes. To achieve time-uncorrelated experiences, 64 simultaneous environments are created, each having a different input reference image. The PPO agent is trained after every five episodes. As for the range of the reward functions, we normalize the reward function so that the values are within $[-1.0, 1.0]$.

While training, the size of the mini batch is set to 32 for four iterative mini epochs. $\gamma$ and $\lambda$, which are used to estimate discounted and generalized advantage, are set to 0.99 and 0.95, respectively. These values encourage the agent to learn long-term dependence of the action choices instead of greedy choices. The learning rate used for training is 1e-4, which shows good training stability. For CNN architectures, we apply weight regularization with 1e-4 as a weight regularization, and gradient normalization (limiting the size of generated gradient while training) is applied with 0.5 for stable training.

These design choices are adjustable depending on the size of the training dataset and problem formulation details. With these architectures and hyperparameter settings, we achieve good performance in training and evaluation results, as presented in the next subsection.

Fig. 14 shows the training loss and reward values during training with the pixel difference reward function. As shown in the figure, the reward saturates around 0.6, and the loss is near 0.002. It saturates near 0.6 because we normalized the rewards between $[-1.0, 1.0]$. The figure shows the stable training of our *GDNet*.

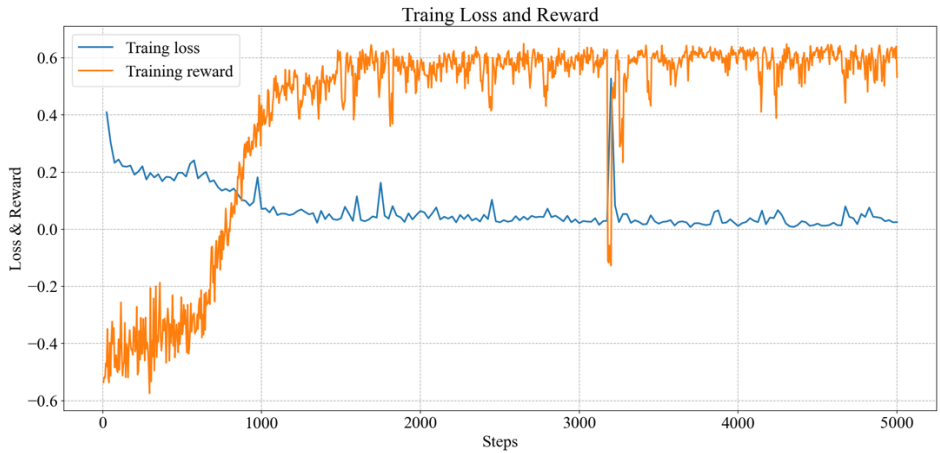

Figure 14. Training loss and reward during training with pixel difference reward.

Additionally, we try the A2C framework (Mnih et al., 2016). However, we cannot observe trained results with many different combinations of hyperparameters. We believe that the clipping-based surrogate loss of the PPO framework contributes heavily to the stable training performance in our problem formulation, which the vanilla A2C cannot achieve.

### 4.3 Experiment Results: Evaluations and Comparisons

We first cross-compare the two rewards, pixel difference and structure dissimilarity, by training an RL agent with each metric and evaluating with both metrics to determine which one achieves more diversity. Then, we compare the result of *GDNet* to equ-distance, greedy, and exhaustive search with a couple of diversity metrics. Lastly, we provide the evaluation results of *GDNet* with unsaturated *TopOpNet* to determine the effects of errors in *TopOpNet* to *GDNet*.

Table 2. Cross-comparison results for the two reward functions.

|  |  | Testing results | |
|---|---|---|---|
|  |  | Pixel difference | Structural dissimilarity |
| Training reward function | Pixel difference | 0.603 | 0.159 |
|  | Structural dissimilarity | 0.426 | 0.130 |

Table 2 shows the cross-comparison results of the two reward functions. While training with each one of the reward functions, we use 500 reference images when training *TopOpNet* and *GDNet* and conduct tests on the remaining 139 reference images for the cross-comparison. Thus, *TopOpNet* and *GDNet* have two sets. Once the networks are trained, their average rewards with both metrics are calculated. As shown in the table, training with pixel difference and testing with pixel difference show the highest score. Training with pixel difference shows a higher score even in structure dissimilarity than training with the dissimilarity. Our experiment results show that pixel difference is a more appropriate rewarding metric in our diverse wheel designing problem because it captures the existence of a material in each pixel, whereas dissimilarity captures the statistical characteristics of designs. Directly comparing pixel by pixel well captures the diversity in our problem.

Next, we compare *GDNet* with equ-distance action, greedy approach, and exhaustive optimal. The equ-distance strategy divides each parameter space as equally as possible. Given that five parameters are selected out of 11 discrete action spaces, four options are available: [0, 2, 5, 8, 10], [0, 3, 5, 7,10], [0, 2, 5, 7, 10], and [0, 3, 5, 8, 10]. Here, the actions are represented with an index among 11 actions. The parameters have physical meanings as described in Section 2. Thus, equal division could possibly lead to diverse designs. While evaluating, we select one with largest gain among the four options.

The greedy strategy selects actions that only consider one-step improvement. In the first step, only a single variation is selected by deciding $c_1$ and $c_2$. Thus, this strategy selects the parameters resulting in the largest pixel difference compared with the reference design. In the next step, deciding another set of $c_1$ and $c_2$ adds three more variations. Similarly, the greedy strategy selects one resulting largest pixel difference among possible candidates. These greedy choices are made for five times, corresponding to an episode in our problem.

Maximum marginal relevance (MMR; Carbonell et al., 1998) is widely used for document classification and retrieval. Given a query, the MMR algorithm reranks searched documents on the basis of its relevance to the query and to the novelty of a document in comparison with the set already selected. In this manner, MMR makes a list of documents relevant to the query while merging duplicated documents. Our problem is similar to the document reranking problem in the sense that it tries to maximize result diversity. However, no relevance term exists in our problem because linking between a reference image and generated designs are the responsibility of topology optimization or *TopOpNet*. Thus, we use the greedy strategy as our comparing strategy.

Lastly, the exhaustive maximum compares all combinations of action sets and selects one with the maximum pixel difference. We have 11 candidates for each parameter; thus, a total of $_{11}C_2 \times _{11}C_2 = 213{,}444$ candidates when duplication is allowed or $11^{10}$ otherwise. In practice, exhaustive search is computationally inefficient and even infeasible when the search space is large. The exhaustive and greedy search approaches must perform a search procedure every time new inputs are injected. On the contrary, *GDNet*, an RL-based search approach, can generalize previously unseen inputs, thus requiring only single inference time.

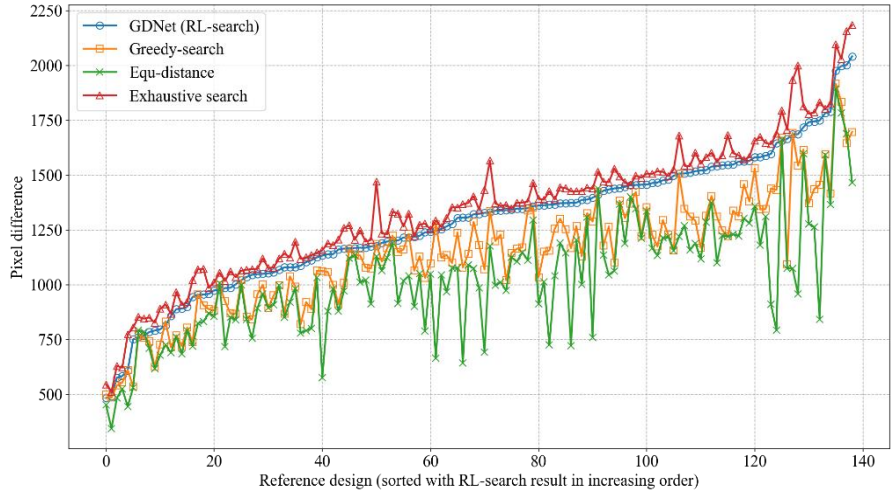

Figure 15. Pixel difference evaluation of *GDNet* with equ-distance, greedy, and exhaustive search on testing reference designs.

Fig. 15 shows the comparison results. The evaluations are performed on testing reference images only. For improved readability, we sort the reference designs in increasing order on the basis of *GDNet*'s result. Pixel difference values are not normalized as in training. As shown in the figure, exhaustive search achieves maximum rewards on the evaluation set. However, as mentioned, exhaustive search needs to compare approximately 213k candidates, thus consuming substantial time. Even with our PC with Intel i7-core and NVIDIA GTX 1080TI GPU and by exploiting GPU parallel computing, comparing all 213k candidates takes approximately 4 h. As expected, equ-distance shows the lowest diversity, and the greedy search achieves the diversity score in between equ-distance and *GDNet*'s results. Given that the greedy algorithm only takes one-step optimal actions, it compromises global optimality.

Table 3. Performance evaluation results.

|  | *GDNet* | Greedy |
|---|---|---|
| Exhaustive search (upper bound) | 95.38% | 87.17% |
| Equ-distance (lower bound) | 140.61% | 113.15% |

Table 3 summarizes the comparison results. The exhaustive search results show the maximum diversity, thus providing the upper bound of the performance. Whereas equ-distance provides the lower bound because it is the most naïve strategy. *GDNet* achieves results close to that of the exhaustive search. The greedy strategy has some performance loss because it only considers one-step optimal, not global optimal. For the same reason, *GDNet* achieves more diverse results from the lower bound compared to the greedy.

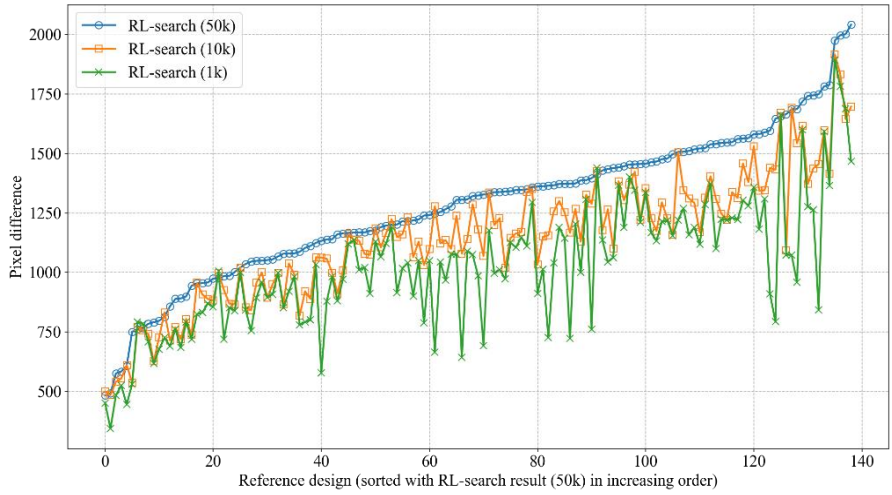

Figure 16. Effects of error on *TopOpNet* to *GDNet*.

Fig. 16 shows the effect of error on *TopOpNet* to *GDNet*'s performance. To determine the effect, three *TopOpNets* are made with the same network architecture and training dataset but with different number in training steps: 1k, 10k, and 50k. As training step increases, *TopOpNet* shows good approximation performance. Thus, the ones with 1k and 10k have errors in approximation. As shown in the figure, errors in topology optimization hugely affect the performance of *GDNet*. This effect is intuitive because *TopOpNet* shows similar results to multiple parameters when training is not completed, leading to an inability to suggest diverse designs.

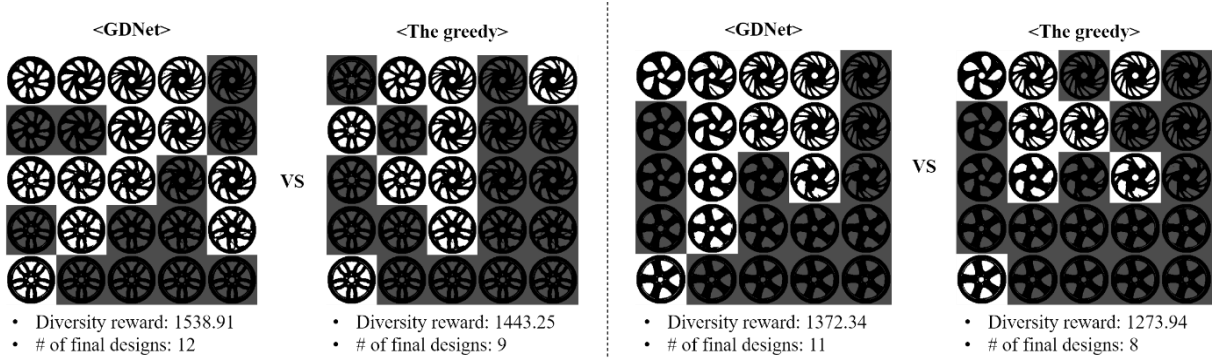

Figure 17. Number of final designs selected with *GDNet* and the greedy strategy.

To visualize the diversity of the variants, we conduct postprocessing on the variants to select a set of final designs as follows: Among the 25 candidates, we first select one that is most different from the reference design in terms of pixel difference. Meanwhile, the candidates are sorted in decreasing order according to pixel difference from the reference. The first one is then added to the selected set. For the next one in the sorted set, if the pixel difference is below a certain threshold to all designs in an already selected set, then it is added to the selected set; otherwise, it is added to the ignored set. We use 97% for the threshold. This process continues until the entire candidates are in either one of the sets. Then, the number of the selected sets are counted to measure roughly the diversity of the generated designs. Fig. 17 shows the selected ones and ignores the one in the shades. As shown in the figure, *GDNet* achieves a higher score in pixel difference, and the candidate set contains more final designs in comparison with the greedy strategy for the two reference numbers. Table 4 summarizes the evaluation results on the number of final designs compared with pixel difference. Intuitively, *GDNet* shows better diversity in pixel difference and the number of final designs. The number of final designs is not directly related to the diversity metric used while training, yet *GDNet* achieves better results in the metric.

Table 4. Comparison of rewards and the number of final designs

|  |  | Diversity reward (pix-diff) | | # of final designs | |
|---|---|---|---|---|---|
|  |  | Mean | Std. | Mean | Std. |
| Algorithm | GDNet | 1281.45 | 304.68 | 9.34 | 2.43 |
|  | The greedy | 1168.48 | 277.43 | 8.83 | 2.16 |

Lastly, we check the trade-off between the diversity (pixel-difference) and engineering performance (compliance) of generated results through Fig. 18. All results are on the Pareto curve. The exhaustive search strategy has the most diversity, and equ-distance has the lowest compliance.

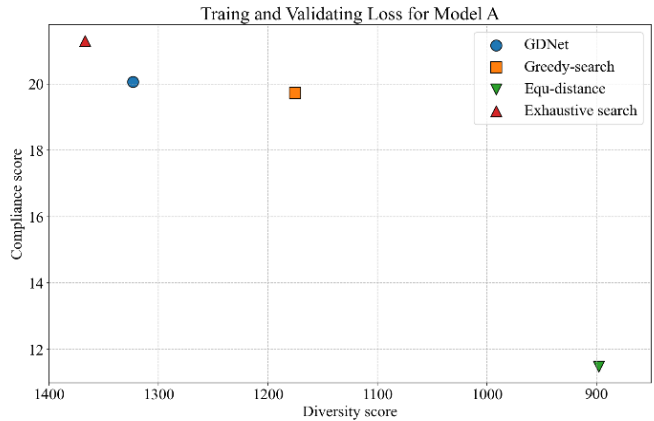

Figure 18. Trade-off between diversity and engineering performance.

In this section, we evaluate the performance of *GDNet* in various ways. In summary, *GDNet*, an RL-based approach, has an advantage in generalization and shows good performance in pixel-difference and log determinant metrics in comparison with other baselines. Given that design suggestion involves images having nonlinear and nontrial features, neural network-based approaches, specifically RL-based ones, show good performance compared with greedy algorithms. We plan to expand our *GDNet* problem with continuous action space in the future.

## 5. Conclusion

This study addresses the generative design problem with a deep neural network. An RL-based generative design framework is proposed to determine the optimal combination of design parameters that can encourage the diversity of generated designs. An automotive wheel designing as a case study demonstrated that the proposed framework finds the optimal design parameter combinations based on given reference designs.

First, we approximated the time-demanding and computation-intensive topology optimization algorithm for the wheel design using a deep neural network, which we call *TopOpNet*. With our proposed network architectures, topology optimization can be computed within an order of milliseconds (an inferencing time), while the algorithm requires an order of 10 min in general. This, in turn, leads to the GPU processing the data, not the CPU. The network architectures have similar components used in semantic segmentation, and we optimized the architectures for our problem.

Second, taking advantage of the reduced computation time of *TopOpNet*, we tackled the maximum generative design diversity problem using the reinforcement framework, *GDNet*. We first found that the problem has non-linear characteristics so that deep learning was a suitable solution. Then, the problem was formulated as a sequential process, suggesting a set of parameters at each step so that the reinforcement framework could be applied. We used VAE regularizing as a mean to accelerate the training process by helping the network extract features. PPO has shown stable training results while others have not. Through extensive experiments, we show that pixel difference is an appropriate metric for our problem, and GDNet obtains better results than the greedy algorithm and the naïve equal-distance strategy.

We are planning to extend the approach to continuous action space, which requires additional computation resources and time. RL-based generative design has the potential to suggest diverse designs, and we plan to extend the approach to other design problems in the future.

**Acknowledgement**

This work was supported by National Research Foundation of Korea (NRF) grants funded by the Korean government [grant numbers 2017R1C1B2005266, 2018R1A5A7025409]. The authors would like to thank Yebin Youn for her help in optimizing the model architecture.

## References

Abueidda, D. W., Koric, S., & Sobh, N. A. (2020). Topology optimization of 2D structures with nonlinearities using deep learning. Computers & Structures, 237, 106283.

Alaimo, G., Auricchio, F., Bianchini, I., & Lanzarone, E. (2018). Applying functional principal components to structural topology optimization. International Journal for Numerical Methods in Engineering, 115(2), 189-208.

Autodesk. (2021). https://www.autodesk.com/solutions/generative-design

Badloe, T., Kim, I., & Rho, J. (2020). Biomimetic ultra-broadband perfect absorbers optimised with reinforcement learning. Physical Chemistry Chemical Physics, 22(4), 2337-2342.

Banga, S., Gehani, H., Bhilare, S., Patel, S., & Kara, L. (2018). 3d topology optimization using convolutional neural networks. arXiv preprint arXiv:1808.07440.

Bendsoe, M. P., & Sigmund, O. (2013). Topology optimization: theory, methods, and applications. Springer Science & Business Media.

Caldas, L. G. (2001). An evolution-based generative design system: using adaptation to shape architectural form (Doctoral dissertation, Massachusetts Institute of Technology).

Cang, R., Yao, H., & Ren, Y. (2019). One-shot generation of near-optimal topology through theory-driven machine learning. Computer-Aided Design, 109, 12-21.

Carbonell, J. & Goldstein, J. (1998). The use of MMR, diversity-based reranking for reordering documents and producing summaries, In Proceeding of the 21$^{st}$ annual international ACM SIGIR conference on research and development in information retrieval (p. 335-336)

Chen, L. C., Papandreou, G., Kokkinos, I., Murphy, K., & Yuille, A. L. (2017). Deeplab: Semantic image segmentation with deep convolutional nets, atrous convolution, and fully connected crfs. IEEE transactions on pattern analysis and machine intelligence, 40(4), 834-848.

Chen, W., & Ahmed, F. (2021). PaDGAN: Learning to Generate High-Quality Novel Designs. Journal of Mechanical Design, 143(3).

Chen, X. A., Tao, Y., Wang, G., Kang, R., Grossman, T., Coros, S., & Hudson, S. E. (2018, April). Forte: User-driven generative design. In Proceedings of the 2018 CHI Conference on Human Factors in Computing Systems (pp. 1-12).

Cui, H., Turan, O., & Sayer, P. (2012). Learning-based ship design optimization approach. Computer-Aided Design, 44(3), 186-195.

Dong, Z., Ho, J., Yu, Y. F., Fu, Y. H., Paniagua-Dominguez, R., Wang, S., ... & Yang, J. K. (2017). Printing beyond sRGB color gamut by mimicking silicon nanostructures in free-space. Nano letters, 17(12), 7620-7628.

Frazer, J. (2002). Creative design and the generative evolutionary paradigm. In Creative evolutionary systems (pp. 253-274). Morgan Kaufmann.

Gao, L., Yang, J., Wu, T., Yuan, Y. J., Fu, H., Lai, Y. K., & Zhang, H. (2019). SDM-NET: Deep generative network for structured deformable mesh. ACM Transactions on Graphics (TOG), 38(6), 1-15.

Guo, T., Lohan, D. J., Cang, R., Ren, M. Y., & Allison, J. T. (2018). An indirect design representation for topology optimization using variational autoencoder and style transfer. In 2018 AIAA/ASCE/AHS/ASC Structures, Structural Dynamics, and Materials Conference (p. 0804).

Ha, D. (2019). Reinforcement learning for improving agent design. Artificial life, 25(4), 352-365.

He, K., Zhang, X., Ren, S., & Sun, J. (2016). Deep residual learning for image recognition. In Proceedings of the IEEE conference on computer vision and pattern recognition (pp. 770-778).

Hoyer, S., Sohl-Dickstein, J., & Greydanus, S. (2019). Neural reparameterization improves structural optimization. arXiv preprint arXiv:1909.04240

Kallioras, N. A., & Lagaros, N. D. (2020). DzAI𝑁: Deep learning based generative design. Procedia Manufacturing, 44, 591-598.

Kazi, R. H., Grossman, T., Cheong, H., Hashemi, A., & Fitzmaurice, G. W. (2017, October). DreamSketch: Early Stage 3D Design Explorations with Sketching and Generative Design. In UIST (Vol. 14, pp. 401-414).

Keshavarzzadeh, V., Kirby, R. M., & Narayan, A. (2020). Stress-based topology optimization under uncertainty via simulation-based Gaussian process. Computer Methods in Applied Mechanics and Engineering, 365, 112992.

Kingma, D. P., & Ba, J. (2014). Adam: A method for stochastic optimization. arXiv preprint arXiv:1412.6980.

Kingma, D. P., & Welling, M. (2013). Auto-encoding variational bayes. arXiv preprint arXiv:1312.6114.

Krish, S. (2011). A practical generative design method. Computer-Aided Design, 43(1), 88-100.

Kumar, T., & Suresh, K. (2019). A density-and-strain-based K-clustering approach to microstructural topology optimization. Structural and Multidisciplinary Optimization, 1-17.

Kunakote, T., & Bureerat, S. (2011). Multi-objective topology optimization using evolutionary algorithms. Engineering Optimization, 43(5), 541-557.

Lee, X. Y., Balu, A., Stoecklein, D., Ganapathysubramanian, B., & Sarkar, S. (2019). A case study of deep reinforcement learning for engineering design: Application to microfluidic devices for flow sculpting. Journal of Mechanical Design, 141(11).

Lei, X., Liu, C., Du, Z., Zhang, W., & Guo, X. (2019). Machine learning-driven real-time topology optimization under moving morphable component-based framework. Journal of Applied Mechanics, 86(1).

Li, B., Huang, C., Li, X., Zheng, S., & Hong, J. (2019). Non-iterative structural topology optimization using deep learning. Computer-Aided Design, 115, 172-180.

Li, J., Xu, K., Chaudhuri, S., Yumer, E., Zhang, H., & Guibas, L. (2017). Grass: Generative recursive autoencoders for shape structures. ACM Transactions on Graphics (TOG), 36(4), 1-14.

Li, K., & Malik, J. (2016). Learning to optimize. arXiv preprint arXiv:1606.01885.

Li, M., Cheng, Z., Jia, G., & Shi, Z. (2019). Dimension reduction and surrogate based topology optimization of periodic structures. Composite Structures, 229, 111385.

Lin, Q., Hong, J., Liu, Z., Li, B., & Wang, J. (2018). Investigation into the topology optimization for conductive heat transfer based on deep learning approach. International Communications in Heat and Mass Transfer, 97, 103-109.

Lindenmayer, A. (1975). Developmental algorithms for multicellular organisms: A survey of l-systems. Journal of Theoretical Biology, 54(1):3–22.

Long, J., Shelhamer, E., & Darrell, T. (2015). Fully convolutional networks for semantic segmentation. In Proceedings of the IEEE conference on computer vision and pattern recognition (pp. 3431-3440).

Liu, K., Tovar, A., Nutwell, E., & Detwiler, D. (2015, August). Towards nonlinear multimaterial topology optimization using unsupervised machine learning and metamodel-based optimization. In International Design Engineering Technical Conferences and Computers and Information in Engineering Conference (Vol. 57083, p. V02BT03A004). American Society of Mechanical Engineers.

Matejka, J., Glueck, M., Bradner, E., Hashemi, A., Grossman, T., & Fitzmaurice, G. (2018, April). Dream lens: Exploration and visualization of large-scale generative design datasets. In Proceedings of the 2018 CHI Conference on Human Factors in Computing Systems (pp. 1-12).

Meinhardt, H. (1976). Morphogenesis of lines and nets. Differentiation, 6(2):117–123.

Mnih, V., Badia, A. P., Mirza, M., Graves, A., Lillicrap, T., Harley, T., ... & Kavukcuoglu, K. (2016, June). Asynchronous methods for deep reinforcement learning. In International conference on machine learning (pp. 1928-1937).

Mnih, V., Kavukcuoglu, K., Silver, D., Graves, A., Antonoglou, I., Wierstra, D., & Riedmiller, M. (2013). Playing atari with deep reinforcement learning. arXiv preprint arXiv:1312.5602.

Mnih, V., Kavukcuoglu, K., Silver, D., Rusu, A. A., Veness, J., Bellemare, M. G., ... & Petersen, S. (2015). Human-level control through deep reinforcement learning. nature, 518(7540), 529-533.

Oh, S., Jung, Y., Lee, I., & Kang, N. (2018, August). Design automation by integrating generative adversarial networks and topology optimization. In International Design Engineering Technical Conferences and Computers and Information in Engineering Conference (Vol. 51753, p. V02AT03A008). American Society of Mechanical Engineers.

Oh, S., Jung, Y., Kim, S., Lee, I., & Kang, N. (2019). Deep generative design: Integration of topology optimization and generative models. Journal of Mechanical Design, 141(11).

Qiu, Zheng, Quhao Li, Shutian Liu, and Rui Xu. "Clustering-based concurrent topology optimization with macrostructure, components, and materials." Structural and Multidisciplinary Optimization (2020): 1-21.

Patel, J., & Choi, S. K. (2012). Classification approach for reliability-based topology optimization using probabilistic neural networks. Structural and Multidisciplinary Optimization, 45(4), 529-543.

Ronneberger, O., Fischer, P., & Brox, T. (2015, October). U-net: Convolutional networks for biomedical image segmentation. In International Conference on Medical image computing and computer-assisted intervention (pp. 234-241). Springer, Cham.

Sajedian, I., Badloe, T., & Rho, J. (2019). Optimisation of colour generation from dielectric nanostructures using reinforcement learning. Optics express, 27(4), 5874-5883.

Sajedian, I., Lee, H., & Rho, J. (2019). Double-deep Q-learning to increase the efficiency of metasurface holograms. Scientific reports, 9(1), 1-8.

Sasaki, H., & Igarashi, H. (2019). Topology optimization accelerated by deep learning. IEEE Transactions on Magnetics, 55(6), 1-5.

Schulman, J., Levine, S., Abbeel, P., Jordan, M., & Moritz, P. (2015, June). Trust region policy optimization. In International conference on machine learning (pp. 1889-1897).

Schulman, J., Moritz, P., Levine, S., Jordan, M., & Abbeel, P. (2015). High-dimensional continuous control using generalized advantage estimation. arXiv preprint arXiv:1506.02438.

Schulman, J., Wolski, F., Dhariwal, P., Radford, A., & Klimov, O. (2017). Proximal policy optimization algorithms. arXiv preprint arXiv:1707.06347.

Shea, K., Aish, R., & Gourtovaia, M. (2005). Towards integrated performance-driven generative design tools. Automation in Construction, 14(2), 253-264.

Singh, V., & Gu, N. (2012). Towards an integrated generative design framework. *Design studies*, *33*(2), 185-207.

So, S., Badloe, T., Noh, J., Bravo-Abad, J., & Rho, J. (2020). Deep learning enabled inverse design in nanophotonics. Nanophotonics, 9(5), 1041-1057.

Sosnovik, I., & Oseledets, I. (2017). Neural networks for topology optimization. arXiv preprint arXiv:1709.09578.

Strömberg, N. (2020). Efficient detailed design optimization of topology optimization concepts by using support vector machines and metamodels. Engineering Optimization, 52(7), 1136-1148.

Sun, H., & Ma, L. (2020). Generative Design by Using Exploration Approaches of Reinforcement Learning in Density-Based Structural Topology Optimization. Designs, 4(2), 10.

Sutskever, I., Vinyals, O., & Le, Q. V. (2014). Sequence to sequence learning with neural networks. In Advances in neural information processing systems (pp. 3104-3112).

Szegedy, C., Ioffe, S., Vanhoucke, V., & Alemi, A. A. (2017, February). Inception-v4, inception-resnet and the impact of residual connections on learning. In Thirty-first AAAI conference on artificial intelligence.

Tire Rack. (2018). https://www.tirerack.com/content/tirerack/desktop/en/wheels.html

Ulu, E., Zhang, R., & Kara, L. B. (2016). A data-driven investigation and estimation of optimal topologies under variable loading configurations. Computer Methods in Biomechanics and Biomedical Engineering: Imaging & Visualization, 4(2), 61-72.

Van Hasselt, H., Guez, A., & Silver, D. (2016, March). Deep reinforcement learning with double q-learning. In Thirtieth AAAI conference on artificial intelligence.

Vlah, D., Žavbi, R., & Vukašinović, N. (2020, May). Evaluation of topology optimization and generative design tools as support for conceptual design. In *Proceedings of the Design Society: DESIGN Conference* (Vol. 1, pp. 451-460). Cambridge University Press.

Wang, L., Tao, S., Zhu, P., & Chen, W. (2020, August). Data-Driven Multiscale Topology Optimization Using Multi-Response Latent Variable Gaussian Process. In International Design Engineering Technical Conferences and Computers and Information in Engineering Conference (Vol. 84003, p. V11AT11A057). American Society of Mechanical Engineers.

Wang, Z., Bovik, A. C., Sheikh, H. R., & Simoncelli, E. P. (2004). Image quality assessment: from error visibility to structural similarity. IEEE transactions on image processing, 13(4), 600-612.

Xiao, M., Chu, S., Gao, L., & Li, H. (2019). A hybrid method for density-related topology optimization. International Journal of Computational Methods, 16(08), 1850116.

Xiao, M., Lu, D., Breitkopf, P., Raghavan, B., Dutta, S., & Zhang, W. (2020). On-the-fly model reduction for large-scale structural topology optimization using principal components analysis. Structural and Multidisciplinary Optimization, 1-22.

Yildiz, A. R., Öztürk, N., Kaya, N., & Öztürk, F. (2003). Integrated optimal topology design and shape optimization using neural networks. Structural and Multidisciplinary Optimization, 25(4), 251-260.

Yonekura, K., & Hattori, H. (2019). Framework for design optimization using deep reinforcement learning. Structural and Multidisciplinary Optimization, 60(4), 1709-1713.

Yoo, S., Lee, S., Kim, S., Hwang, K. H., Park, J. H., & Kang, N. (2020). Integrating Deep Learning into CAD/CAE System: Case Study on Road Wheel Design Automation. arXiv preprint arXiv:2006.02138.

Yu, Y., Hur, T., Jung, J., & Jang, I. G. (2019). Deep learning for determining a near-optimal topological design without any iteration. Structural and Multidisciplinary Optimization, 59(3), 787-799.

Zhang, Y., Chen, A., Peng, B., Zhou, X., & Wang, D. (2019). A deep Convolutional Neural Network for topology optimization with strong generalization ability. arXiv preprint arXiv:1901.07761.

Zhou, Y., & Saitou, K. (2017). Topology optimization of composite structures with data-driven resin filling time manufacturing constraint. Structural and Multidisciplinary Optimization, 55(6), 2073-2086.

# Appendix

In appendix, we present another TopOpNet architecture we have searched while approximating topology optimization process followed by another evaluation result comparing GDNet, the greedy, equ-distance, and exhaustive in logdet metric.

## 1. Alternative *TopOpNet* Architecture

The second model we use for *TopOpNet*, model B for simplicity, reduces the number of trainable parameters to boost the speed of training while slightly sacrificing performance. Fig. 19 shows the overall architecture of model B. Like the previous model, the general architecture consists of encoding and decoding blocks. However, only one static block was used for each part in both the encoding and decoding blocks.

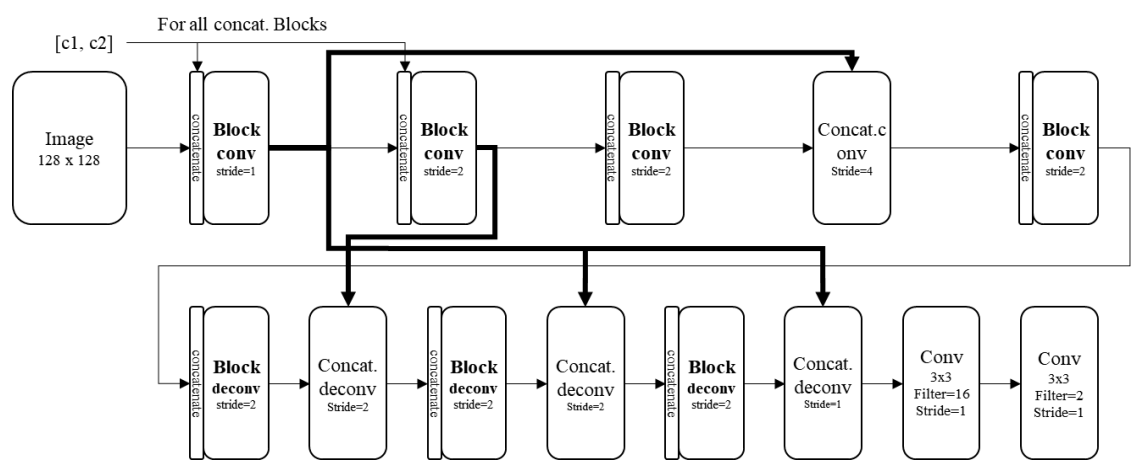

Figure 19. Network architecture with customized skip layers (Model B).

Fig. 20 shows the encoding and decoding blocks used in model B. The encoding block decreases the size of the input tensor by $s$ using the stride size. It has a main stream of layers with a residual connection. Conversely, the decoding block has mainly deconvolution layers and enlarges the size of the input tensor by $s$ using strides.

We have come up with this architecture by trying out all possible combinations of skip layers. For a skip layer, the shape of input and output tensor layer needs to be the same. The performance in many metrics is measured for each candidate and the best one is chosen as our reference architecture. But the gaps between the performance are not significant compared to that between *TopOpNet* in the manuscript and the alternative model. So, we omit the comparison results.

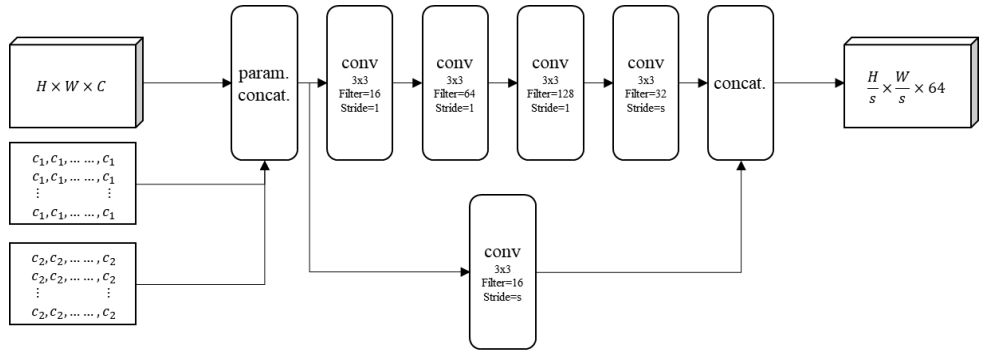

(a) Encoding block

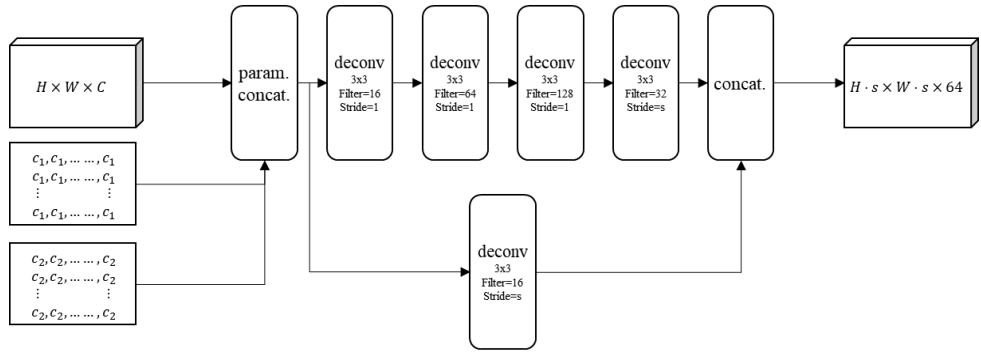

(a) Decoding block

Figure 20. Encoding and decoding blocks in model B.

The number of parameters for this model was approximately 1.25 million, which is 1/10 of that of *TopOpNet*. Because our problem, approximating the topology optimization algorithm, is somewhat restricted due to the limited amount of data, this simplified model was considered as having good performance.

## 2. Evaluation of *GDNet* in logdet

Besides the two rewards, we also consider gaussian kernel based *logdet* diversity measurement as Eq. (19)-(20). The determinant of a $n \times n$ matrix measures $n$ dimensional volume created by $n$ transformed vectors by the matrix from unit vectors in original coordinate space. The volume increases (decreases) as the matrix have many (less) independent components, meaning more (less) diversity. Here, the matrix is composed of a Gaussian kernel with pixel difference between two designs. However, we find that the determinant-based metric is not suitable for our problem because the determinant requires certain criterions to exist. In other words, the determinant value does not always exist, therefore infeasible for our problem formulation. For this reason, we limit the use of the gaussian kernel *logdet* diversity metric as an evaluating metric, not as a rewarding metric in training.

$$logdet(A) = \log(\det(A)) \tag{19}$$

$$A = \left[-0.5 \cdot \exp(x_i - x_j)^2\right] \; for \; all \; i,j \tag{20}$$

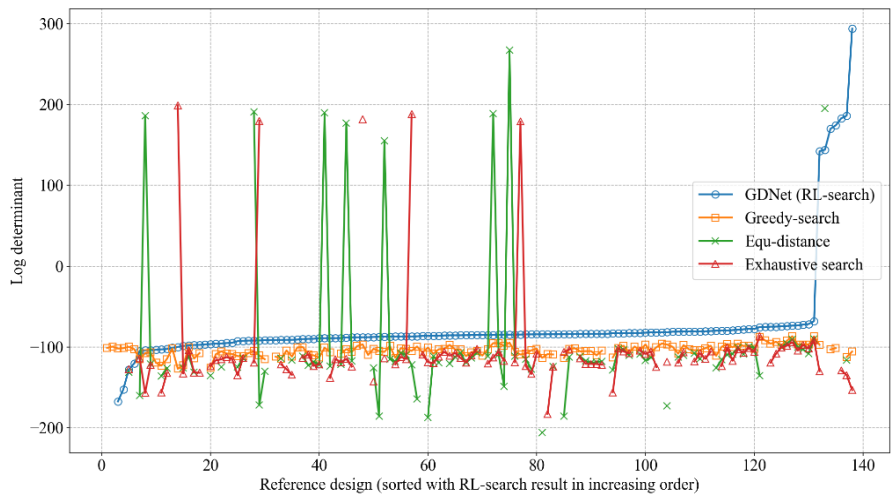

Figure 21. Logdet evaluation of *GDNet* with equ-distance, greedy, and exhaustive search on testing reference designs.

Fig. 21 shows the same comparison with diversity score (*logdet*) metric proposed by (Chen & Ahmed, 2020). As states, there might not be a determinant for some matrix, so, there are some missing points in the graph. Like the previous graph, the reference designs are sorted in increasing order with the results of *GDNet*. There are some designs that equ-distance show extremely better log determinant values than our *GDNet*. However, for most of the cases, equ-distance and the greedy show lower log determinants than *GDNet*. Exhaustive results also show relatively low log determinants than ours. This instability originates from the characteristics of log determinant. Our log determinant is based on pixel difference and involves exponential scaling functions, *exp* and *log*, the value sometimes explodes which leads to the instable results.